\documentclass[journal]{IEEEtran}

\usepackage{multirow,booktabs,color,soul,threeparttable,rotating}
\usepackage{amsmath,subfigure}
\usepackage{epsfig,amssymb}
\usepackage{algorithmic}
\usepackage{algorithm}

\graphicspath{{Figure/}}
\usepackage{url}
\definecolor{hl}{rgb}{0.75,0.75,0.75}
\sethlcolor{hl}

\begin{document}

\title{Evolutionary Multiobjective Optimization Driven by Generative Adversarial Networks (GANs)}

\author{Cheng~He~\textit{IEEE Member},
                Shihua~Huang,
                Ran~Cheng~\textit{IEEE Member},\\
				Kay Chen Tan \textit{IEEE Fellow},
                and Yaochu Jin \textit{IEEE Fellow}
\thanks{C. He, S. Huang, and R. Cheng are with the University Key Laboratory of Evolving Intelligent Systems of Guangdong Province, Department of Computer Science and Engineering, Southern University of Science and Technology, Shenzhen 518055, China.
E-mail: chenghehust@gmail.com, shihuahuang95@gmail.com, ranchengcn@gmail.com. (\emph{Corresponding author: Ran Cheng})}
\thanks{K. C. Tan is with the Department of Computer Science, City University of Hong Kong, Hong Kong. E-mail: kaytan@cityu.edu.hk.}
\thanks{Y. Jin is with the Department of Computer Science, University of Surrey, Guildford, Surrey, GU2 7XH, United Kingdom. Email: yaochu.jin@surrey.ac.uk.}
\thanks{This work was supported in part by the National Natural Science Foundation of China (No. 61903178 and 61906081), in part by the Program for Guangdong Introducing Innovative and Entrepreneurial Teams grant (No. 2017ZT07X386), in part by the Shenzhen Peacock Plan grant (No. KQTD2016112514355531), in part by the Program for University Key Laboratory of Guangdong Province grant (No. 2017KSYS008), and in part by the Research Grants Council of the Hong Kong SAR (No. CityU11202418 and CityU11209219).}
}

\markboth{IEEE Transactions on Cybernetics,~Vol.~, No.~, month~year}%
{He \MakeLowercase{\textit{et al.}}: Evolutionary Multiobjective Optimization Driven by Generative Adversarial Networks (GANs)}

\maketitle

\begin{abstract}
Recently, increasing works have proposed to drive evolutionary algorithms using machine learning models.
Usually, the performance of such model based evolutionary algorithms is highly dependent on the training qualities of the adopted models.
Since it usually requires a certain amount of data (i.e., the candidate solutions generated by the algorithms) for model training, the performance deteriorates rapidly with the increase of the problem scales, due to the curse of dimensionality.
To address this issue, we propose a multiobjective evolutionary algorithm driven by the generative adversarial networks (GANs).
At each generation of the proposed algorithm, the parent solutions are first classified into \emph{real} and \emph{fake} samples to train the GANs; then the offspring solutions are sampled by the trained GANs.
Thanks to the powerful generative ability of the GANs, our proposed algorithm is capable of generating promising offspring solutions in high-dimensional decision space with limited training data.
The proposed algorithm is tested on 10 benchmark problems with up to 200 decision variables.
Experimental results on these test problems demonstrate the effectiveness of the proposed algorithm.
\end{abstract}


\section{Introduction}\label{sec:introduction}

Multiobjective optimization problems (MOPs) refer to the optimization problems with multiple conflicting objectives \cite{app-network}, e.g., structure learning for deep neural networks \cite{liu2018structure}, energy efficiency in building design \cite{app-build}, and cognitive space communication \cite{ferreira2017multi}.
The mathematical formulation of the MOPs is presented as follows~\cite{deb2014multi}:
\begin{eqnarray}\label{eq:MOP}
\text{Minimize}& F(\mathbf{\mathbf{x}})=\!(f_1(\mathbf{x}),f_2(\mathbf{x}),\dots,f_M(\mathbf{x}))&\\
\text{subject to}&\mathbf{x}\in X, \nonumber
\end{eqnarray}
where $X$ is the search space of decision variables, $M$ is the number of objectives, and $\mathbf{x}$$=$$(x_1,\dots,x_D)$ is the decision vector with $D$ denoting the number of decision variables~\cite{tian2017effectiveness}.

Different from the single-objective optimization problems with single global optima, there exist multiple optima that trade off between different conflicting objectives in an MOP~\cite{PD}.
In multiobjective optimization, the Pareto dominance relationship is usually adopted to distinguish the qualities of two different solutions~\cite{ENS}.
A solution $\mathbf{x}_A$ is said to Pareto dominate anther solution $\mathbf{x}_B$ ($\mathbf{x}_A\prec \mathbf{x}_B$) \textit{iff}
\begin{equation}
\left\{
\begin{array}{lr}
 \forall i\in 1,2,\dots,M, f_i(\mathbf{x}_A) \leq  f_i(\mathbf{x}_B),\\
 \exists j\in 1,2,\dots,M, f_j(\mathbf{x}_A) <  f_j(\mathbf{x}_B).
\end{array}
\right.
\end{equation}
The collection of all the Pareto optimal solutions in the decision space is called the Pareto optimal set (PS), and the projection of the PS in the objective space is called the Pareto optimal front (PF).
The goal of multiobjective optimization is to obtain a set of solutions for approximating the PF in terms of both convergence and diversity, where each solution should be close to the PF and the entire set should be evenly spread over the PF.

To solve MOPs, a variety of multiobjective evolutionary algorithms (MOEAs) have been proposed, which can be roughly classified into three categories~\cite{RVEA}: the dominance-based algorithms (e.g., the elitist non-dominated sorting genetic algorithm (NSGA-II) \cite{NSGA-II} and the improved strength Pareto EA (SPEA2)~\cite{SPEA2}); the decomposition-based MOEAs (e.g., the MOEA/D~\cite{MOEAD} and MOEA/D using differential evolution (MOEA/D-DE)~\cite{MOEADDE}); and the performance indicator-based algorithms (e.g., the $\mathcal{S}$-metric selection based MOEA (SMS-EMOA) \cite{SMSEMOA} and the indicator based EA (IBEA)~\cite{IBEA}).
There are also some MOEAs not falling into the three categories, such as the third generation differential evolution algorithm (GDE3)~\cite{GDE3}, the memetic Pareto achieved evolution strategy (M-PAES)~\cite{knowles2000m}, and the two-archive based MOEA (Two-Arc) ~\cite{praditwong2006new}, etc.

 \begin{figure}[!htbp]
 \centering
  \includegraphics[width=0.68\linewidth]{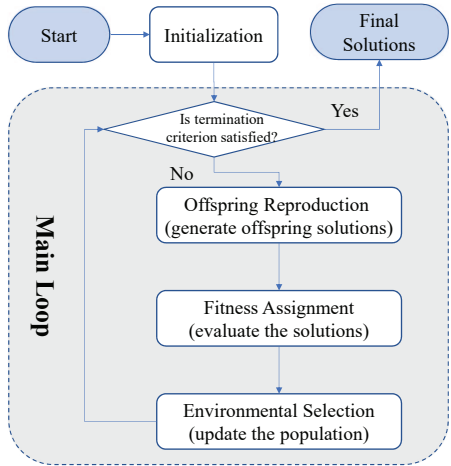}
  \caption{The general framework of MOEAs.}\label{fig:EA}
\end{figure}

\begin{figure}[!htbp]
\centering
  \includegraphics[width=0.7\linewidth]{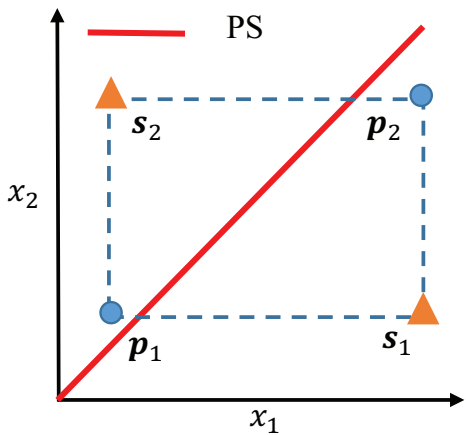}
\caption{An example of the genetic operator (SBX~\cite{PM}) based offspring generation in a 2-D decision space, where $\mathbf{p}_1, \mathbf{p}_2$ denote the parent solutions, and $\mathbf{s}_1, \mathbf{s}_2$ denote the offspring solutions.}\label{fig:rotate}
\end{figure}

In spite of the various technical details adopted in different MOEAs, most of them share a common framework as displayed in Fig. \ref{fig:EA}.
Each generation in the main loop of the MOEAs consists of three operations: offspring reproduction,  fitness assignment, and environmental selection~\cite{eiben2015evolutionary}.
To be specific, the algorithms start from the population initialization; then the offspring reproduction operation will generate offspring solutions; afterwards, the generated offspring solutions are evaluated using the real objective functions; finally, the environmental selection will select some high-quality candidate solutions to survive as the population of the next generation.
In conventional MOEAs, since the reproduction operations are usually based on stochastic mechanisms (e.g., crossover or mutation), the algorithms are unable to explicitly learn from the environments (i.e., the fitness landscapes).
For instance, conventional EAs use the mating selection strategy to select some promising parent solutions based on their fitness values, and then randomly crossover two of them to generate offspring solutions.
For conventional crossover operators such as SBX~\cite{PM},  the offspring solutions will distribute around the vertices of a hyper-rectangle in parallel with the axes of decision variables, and its longest diagonal is the line segment of the two chosen parent solutions.
If the PS of an MOP is not parallel with any axis of decision variable, especially when the PS has a 45$^\circ$ angle to all of the axes (e.g., IMF1 to IMF3 problems in~\cite{IM-MOEA}), there is only a little chance that the offspring solutions will fall around the PS, resulting in the inefficiency of conventional crossover in offspring generation.
An example of the SBX based offspring generation in a 2-D decision space is given in Fig.~\ref{fig:rotate}, where the generated offspring solutions $\mathbf{s}_1, \mathbf{s}_2$ are far from their parents $\mathbf{p}_1,\mathbf{p}_2$ and the PS.

To address the above issue, a number of recent works have been dedicated to designing EAs with learning ability, known as the model based evolutionary algorithms (MBEAs)~\cite{MBEA,zhang2011evolutionary}.
The basic idea of MBEAs is to replace the heuristic operations or the objective functions with computationally efficient machine learning models, where the candidate solutions sampled from the population are used as training data.
Generally, the models are used for the following three main purposes when adopted in MOEAs.

First, the models are used to approximate the real objective functions of the MOP during the fitness assignment process.
MBEAs of this type are also known as the surrogate-assisted EAs~\cite{Jin2000On}, which use computationally cheap machine learning models to approximate the computationally expensive objective functions~\cite{jin2009systems}.
They aim to solve computationally expensive MOPs using a few real objective function evaluations as possible~\cite{jin2011review,SA2017}.
A number of surrogate-assisted MOEAs were proposed in the past decades, e.g., the $S$-metric selection-based EA (SMS-EGO)~\cite{SMS-EGO}, the Pareto rank learning based MOEA~\cite{seah2012pareto}, and the MOEA/D with Gaussian process (GP)~\cite{GP} (MOEA/D-EGO)~\cite{MOEADEGO}.

Second, the models are used to predict the dominance relationship \cite{ParetoSVM} or the ranking of candidate solutions \cite{lu2012classification,bhatt2015novel} during the reproduction or environmental selection process.
For example, in the classification based pre-selection MOEA (CPS-MOEA)~\cite{CPSMOEA}, a k-nearest neighbor (KNN)~\cite{KNN} model is adopted to classify the candidate solutions into \textit{positive} and \textit{negative} classes.
Then the \textit{positive} candidate solutions are selected to survival~\cite{zhang2018preselection}.
Similarly, the classification based surrogate-assisted EA (CSEA) used a feedforward neural network~\cite{svozil1997introduction} to predict the dominance classes of the candidate solutions in evolutionary multiobjective optimization~\cite{CSEA}.

Third, the models are used to generate promising candidate solutions during the offspring reproduction process.
The MBEAs of this type mainly include the multiobjective estimation of distribution algorithms (MEDAs)~\cite{EDA} as well as the inverse modeling based algorithms~\cite{giagkiozis2014pareto}.
The MEDAs estimate the distribution of promising candidate solutions by training and sampling models in the decision space~\cite{karshenas2013multiobjective}.
Instead of generating offspring solutions via crossover or mutation from the parent solutions, the MEDAs explore the decision space of potential solutions by building and sampling explicit probabilistic models of the promising candidate solutions~\cite{eda-initial,sun2018improved}.
Typical algorithms include the Bayesian multiobjective optimization algorithm (BMOA)~\cite{BMOA}, the naive mixture-based multiobjective iterated density estimation EA (MIDEA)~\cite{MIDEA}, the multiobjective Bayesian optimization algorithm (mBOA)~\cite{mBOA}, and the regularity model based MEDA (RM-MEDA)~\cite{RM-MEDA}, etc.
For example, in the covariance matrix adaptation based MOEA/D (MOEA/D-CMA)~\cite{MOEADCMA}, the covariance matrix adaptation model~\cite{loshchilov2013cma} is adopted for offspring reproduction.
As for the inverse modeling based algorithms, they sample points in the objective space and then build inverse models to map them back to the decision space, e.g., the Pareto front estimation method~\cite{giagkiozis2014pareto}, the Pareto-adaptive $\epsilon$-dominance-based algorithm ($pa\lambda$-MyDE)~\cite{hernandez2007pareto}, the reference indicator-based MOEA (RIB-EMOA)~\cite{martinez2014using}, and the MOEA using GP based inverse modeling (IM-MOEA)~\cite{IM-MOEA}.

Despite that existing MBEAs have shown promising performance on a number of MOPs, their performance deteriorates rapidly as the number of decision variables increases.
There are mainly two difficulties when applying existing MBEAs to multiobjective optimization.
First, the requirement of training data for building and updating the machine learning models increases exponentially as the number of decision variables becomes larger, i.e., the MBEAs severely suffer from the curse of dimensionality~ \cite{cd,wang2015memetic}.
Second, since there are multiple objectives involved in MOPs, it is computationally expensive to employ multiple models for sampling different objectives.

The generative adversarial networks (GANs) are generative models that have been successfully applied in many areas, e.g., image generation~\cite{GAN}, unsupervised representation learning~\cite{radford2015unsupervised}, and image super-resolution~\cite{ledig2017photo}.
They are capable of learning the regression distribution over the given/target data in an adversarial manner.
Meanwhile, the candidate solutions can be seen as samples by the distribution of the PS in evolutionary multiobjective optimization.
Under mild conditions,  a PS is an ($M$$-$1)-dimensional manifold, given that $M$ is the number of the objectives~\cite{IM-MOEA}.
Hence, there are two main motivations of using GANs for reproduction in evolutionary multiobjective optimization.
First, it is intuitive to sample candidate solutions using GANs for the estimation of the distribution of the solution set in multiobjective optimization.
Second, it is a natural character of GANs that the samples can be divided into fake and real ones, which is somehow consistent with the nature that the candidate solutions can be divided in multiobjective optimization  (i.e., dominated and non-dominated solutions).
Furthermore, it is naturally suitable to drive evolutionary multiobjective optimization using GANs due to the following reasons.
First, the pairwise generator and discriminator in GANs are capable of distinguishing and sampling promising candidate solutions, which is particularly useful in multiobjective optimization in terms of the Pareto dominance relationship.
Second, thanks to the adversarial learning mechanism, the GANs are able to learn high-dimensional distributions efficiently with limited training data.
By taking such advantages of GANs, we propose a GAN-based MOEA, termed GMOEA.
To the best of our knowledge, it is the first time that the GANs are used for driving evolutionary multiobjective optimization.
The main new contributions of this work can be summarized as follows:
\begin{enumerate}
  \item
    In contrast to conventional MBEAs which are merely dependent on given data (i.e., the candidate solutions), the GANs are able to reuse the data generated by themselves.
    To take such an advantage, in GMOEA,  we propose a classification strategy to classify the candidate solutions into \textit{real} and \textit{fake} data points which are reused as training data.
    This is particularly meaningful for data augmentation in high-dimensional decision space.
  \item
	We sample a multivariate normal Gaussian distribution as the input of GANs in the proposed GMOEA.
	Specifically, the distribution is learned from the promising candidate solutions which approximate the non-dominated front obtained at each generation.
\end{enumerate}

The rest of this paper is organized as follows.
In Section \ref{sec:related}, we briefly review the background of the GANs and other related works.
The details of the proposed GMOEA are presented in Section \ref{sec:algorithm}.
Experimental settings and comparisons of GMOEA with the state-of-the-art MOEAs on the benchmark problems are presented in Section \ref{sec:result}.
Finally, conclusions are drawn in Section \ref{sec:conclusion}.

\section{Background}\label{sec:related}
\subsection{Generative Adversarial Networks}

The generative adversarial networks have achieved considerable success as a framework of generative models~\cite{GAN}.
In general, the GANs produce a model distribution $P_{\hat{\mathbf{x}}}$ (i.e., the distribution of the fake/generated data) that mimics a target distribution $P_{\mathbf{x}}$ (i.e., the distribution of the real/given data).

A pair of GANs consist of a generator and a discriminator, where the generator maps Gaussian noise $\mathbf{z}$ ($\mathbf{z}\in P_{\mathbf{z}}$) to a model distribution $G(\mathbf{z})$  and the discriminator outputs probability $D(\mathbf{x})$ with $\mathbf{x} \in P_{\mathbf{x}}$ $\bigwedge$ $\mathbf{x} \notin P_{\hat{\mathbf{x}}}$.
Generally speaking, the discriminator seeks to maximize probability $D(\mathbf{x})$ ($\mathbf{x}\in P_{\mathbf{x}}$) and minimize probability $D(G(\mathbf{z}))$, while the generator aims to generate more realistic samples to maximize probability $D(G(\mathbf{z}))$, trying to cheat the discriminator.
To be more specific, those two networks are trained in an adversarial manner using the min-max value function $V$:
\begin{eqnarray}\label{eq:GAN}
    &\min\limits_{G} \max\limits_{D}V(D, G) =\\\nonumber
    &\mathbb{E}_{\mathbf{x}\in{P_{\mathbf{x}}}}[logD(\mathbf{x})] + \mathbb{E}_{\mathbf{z}  \in {P_{\mathbf{z}}}}[log(1-D(G(\mathbf{z})))].
\end{eqnarray}

\begin{algorithm}[h]
\caption{Training of the GANs}
\label{al:miniBP}
\begin{algorithmic}[1]
\REQUIRE ~~\\
$P_{\mathbf{x}}$ (given data), $P_{\mathbf{z}}$ (Gaussian noise), $m$ (batch size).
\FOR{total number of training iterations}
        \STATE $\mathbf{X}'\leftarrow P_{\mathbf{x}}$
\FOR{$i\leftarrow 1 : |P_{\mathbf{x}}|/m$}
    \STATE /***** Update the discriminator ****/
        \STATE $\mathbf{T}\leftarrow$ Randomly sample $m$ data points from $\mathbf{X}'$
        \STATE $\mathbf{X'}\leftarrow \mathbf{X'}\backslash \mathbf{T}$
\STATE $\mathbf{Z}\leftarrow$ Sample $m$ noise data points from $P_{\mathbf{z}}$
{\color{black}\STATE Update the discriminator according to (\ref{eq:dis2}) by using $\mathbf{T}$ and $\mathbf{Z}$
    \STATE /****** Update the generator ******/
        \STATE $\mathbf{Z}\leftarrow$ Sample $m$ noise data points from $P_{\mathbf{z}}$
        \STATE Update the generator according to (\ref{eq:gen2})} by using $\mathbf{Z}$
\ENDFOR
\ENDFOR
\end{algorithmic}
\end{algorithm}

Algorithm \ref{al:miniBP} presents the detailed procedures of the training process.
First, $m$ samples are sampled from a Gaussian distribution and the given data (target distribution), respectively.
Second, the discriminator is updated using the gradient descending method according to:
\begin{equation}\label{eq:dis2}
\bigtriangledown \theta_d \frac{1}{m} \sum_{i=1}^m[logD(\mathbf{x}_i) + log\left(1-D(G(\mathbf{z}_i))\right)].
\end{equation}
Sequentially, the generator is updated using the gradient descending method according to:
\begin{equation}\label{eq:gen2}
\bigtriangledown \theta_g \frac{1}{m} \sum_{i=1}^m[log\left(1-D(G(\mathbf{z}_i))\right),
\end{equation}
 where $\mathbf{z}_i$ is a vector randomly sampled from a Gaussian distribution.
 The above procedures are repeated for a number of iterations~\cite{kingma2014adam}.

\subsection{Improved Strength Pareto Based Selection}\label{sec:spea2}

The improved strength Pareto based EA (SPEA2)~\cite{SPEA2} is improved from its original version (SPEA)~\cite{SPEA} by incorporating a tailored fitness assignment strategy, a density estimation technique, and an enhanced truncation method.
In the tailored fitness assignment strategy, the dominance relationship between the pairwise candidate solutions is first detected, and then a strength value is assigned to each candidate solution.
This value indicates the number of candidate solutions it dominates:
\begin{equation}\label{eq:dominance}
Str(\mathbf{x}_i)=|\{j| j\in P\wedge \mathbf{x}_i \prec \mathbf{x}_j \}|,
\end{equation}
where $P$ is the population and $\mathbf{x}_i,\mathbf{x}_j$ are the candidate solutions in it.
Besides, the raw fitness can be obtained as:
\begin{equation}\label{eq:rf}
Raw(\mathbf{x}_i)=\sum^N_{j\in P \wedge \mathbf{x}_j \prec \mathbf{x}_i}{Str(\mathbf{x}_j)}.
\end{equation}
Moreover, the additional density information, termed $Den$, is used to discriminate the candidate solutions having identical raw fitness values.
The density of a candidate solution is defined as:
\begin{equation}\label{eq:df}
Den(\mathbf{x}_i)=\frac{1}{\sigma^k_i+2},
\end{equation}
where $k$ is the square root of the population size, and $\sigma^k_i$ denotes the $k$th nearest Euclidean distance from $\mathbf{x}_i$ to the candidate solutions in the population.
Finally, the fitness can be calculated as
\begin{equation}\label{eq:fit}
Fit(\mathbf{x}_i)= Raw(\mathbf{x}_i)+Den(\mathbf{x}_i).
\end{equation}

The environmental selection of SPEA2 aims to select $N$ solutions from population $P$.
It first selects all the candidate solutions with $Fit$$<$$1$ into set $A$.
If the size of $A$ is smaller than $N$, $N$ solutions with the best $Fit$ are selected from $P$; otherwise, a truncation procedure is invoked to iteratively remove candidate solutions from $A$ until its size equals to $N$, where the candidate solution with the minimum Euclidean distance to the solutions in $A$ is removed each time.

Since the density information is well used, the environmental selection in SPEA2 maintains a set of diverse candidate solutions.
In this work, we adopt it for solution classification and environmental selection in our proposed GMOEA, where the details will be presented in Section \ref{sec:algorithm}.B.

\section{The Proposed Algorithm}\label{sec:algorithm}

The main scheme of the proposed GMOEA is presented in Algorithm \ref{al:framework}.
First, a population $P$ of size $N$ and a pair of GANs are randomly initialized, respectively.
Then the candidate solutions in $P$ are classified into two different datasets with equal size (labeled as \textit{fake} and \textit{real}) and used to train the GANs.
Next, a  set $Q$ of $N$ offspring solutions is generated by the proposed hybrid reproduction strategy.
Afterwards, $N$ candidate solutions are selected from the combination of $P$ and $Q$ by environmental selection.
Finally, the solution classification, model training, offspring reproduction, and environmental selection are repeated until the termination criterion is satisfied.
We will not enter the details of the environmental selection as it is similar to the solution classification, except that the environmental selection takes $N$ solutions from the combination of $P$ and $Q$ as input (instead of selecting half of the solutions from $P$ only) and only outputs the \textit{real} solutions.

\begin{algorithm}[h]
\caption{General Framework of GMOEA}
\label{al:framework}
\begin{algorithmic}[1]
\REQUIRE ~~\\
$N$ (population size),  $m$ (batch size)
\STATE   $P\leftarrow$ Initialize a population of size $N$
\STATE  $GAN\leftarrow$ Initialize the GANs
\WHILE{termination criterion not fulfilled}
    \STATE $\mathbf{X}\leftarrow \text{Solution Classification}$ /*Half of the solutions in $P$ are classified as \textit{fake} samples*/
    \STATE $\mathbf{net} \leftarrow \text{Model Training}$ {/*Use $\mathbf{X}$ to train the model*/}
    \STATE $Q\leftarrow \text{Offspring Reproduction}$ {/*Generate $N$ offspring solutions by the proposed reproduction method*/}
    \STATE $P\leftarrow \text{Environmental Selection}$ {/*Select $N$ solutions from the combination of $P$ and $Q$*/}
\ENDWHILE
\RETURN $P$
\end{algorithmic}
\end{algorithm}

\subsection{Solution Classification}

Solution classification is used to divide the population into two different datasets (\textit{real} and \textit{fake}) for training the GANs.
The \textit{real} solutions are those better-converged and evenly distributed candidate solutions; by contrast, the \textit{fake} ones are those of relatively poor qualities.
We use the environmental selection strategy as introduced in Section \ref{sec:spea2} to select half of the candidate solutions in the current population as \textit{real} samples and the rest as \textit{fake} ones.

\begin{algorithm}[!htbp]
\caption{Solution Classification}
\label{al:select}
\begin{algorithmic}[1]
\REQUIRE
$N'$ (number of fake samples), $P$ (population).
\STATE $Fit\leftarrow$ Calculate the fitness values of candidate solutions in $P$ according to (\ref{eq:fit})
\STATE $A\leftarrow \arg\limits_{\mathbf{x}_i\in P}{Fit(\mathbf{x}_i)<1}$
\IF{$|A|\leq N'$ }
    \STATE  {{$A$}}$\leftarrow$ Select $N'$ candidate solutions with the minimal $Fit$
    \ELSE
    \WHILE{$A>N'$}
    \STATE Delete $\arg\min\limits_{\mathbf{x}_j\in A}{\min{dis(\mathbf{x}_j, A\backslash \mathbf{x}_j)}}$ in $A$
    \ENDWHILE
\ENDIF
\STATE $A\leftarrow$ \textit{real}
\STATE $P\backslash A\leftarrow$ \textit{fake}
\RETURN {$\mathbf{X}\leftarrow \{A\bigcup (P\backslash A), \{\textit{real}\}^{\mathbb{N'}\times 1}\bigcup\{\textit{fake}\}^{\mathbb{N'}\times 1}\}$/*$\mathbf{X}$ is a tuple, where the first element denotes the decision vector and the second one denotes the label*/}
\end{algorithmic}
\end{algorithm}

The pseudo codes of the solution classification are presented in Algorithm \ref{al:select}.
Generally, the purpose of solution classification is to select a set of high-quality candidate solutions in terms of convergence and diversity.
The first term is intuitive, which aims to enhance the selection pressure for pushing the population towards the PF.
The second term aims to satisfy the identity independent distribution assumption for better generalization of the GANs~\cite{DL}.

\subsection{Model Training}

\begin{figure}[!htbp]
  \centering
  \includegraphics[width=\linewidth]{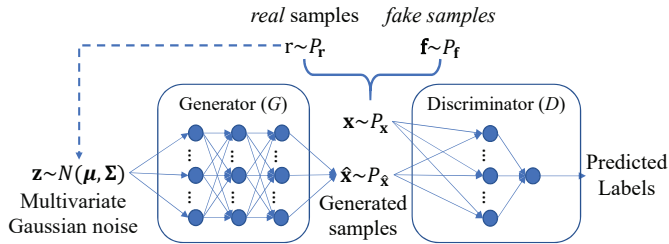}
  \caption{The general scheme of model training in the proposed GMOEA.}\label{fig:GANs}
\end{figure}

The structures of the generator and discriminator adopted in this work are feedforward neural networks~\cite{ANN} with two hidden layers and one hidden layer (the number of neurons in each layer is $D$), respectively.
{Here we adopt this simple structure for the following two reasons.
First, if a more powerful model is adopted, the amount of training data should be increased, resulting in the rapid increase in the computational cost in terms of both CPU time and the number of function evaluations.
For the MOPs we try to solve in this work, the current model is good enough.
Second, in the area of evolutionary optimization, similar simple networks have also been validated in other recent works, e.g. ~\cite{CSEA,zhang2018computational}.
We surmise that it is due to the fact that the problem scales in evolutionary optimization are much smaller than those in other applications such as image processing, such that a simple network will work properly.}
The general scheme of the GANs is given in Fig.~\ref{fig:GANs}, where the distributions of the \textit{real} and \textit{fake} datasets are denoted as $P_{\mathbf{r}}$ and $P_{\mathbf{f}}$, respectively.
The activation functions of the output layers in these two networks are sigmoid functions to ensure that the output values vary in $[0, 1]$.
Here, we propose a novel training method to take advantage of the labeled samples.

First, the mean vector and covariance matrix of the \textit{real} samples are calculated by:
\begin{equation}\label{eq:mean}
\begin{aligned}
\boldsymbol{\mu}&= \frac{\sum_{i=1}^{\lfloor N/2\rfloor}{\mathbf{r}_i}}{\lfloor N/2\rfloor},\\
\boldsymbol{\Sigma} &= \frac{\sum_{i=1}^{\lfloor N/2\rfloor}{(\mathbf{r}_i-\boldsymbol{\mu})(\mathbf{r}_i}-\boldsymbol{\mu})^\text{T}}{\lfloor N/2\rfloor-1},\\
\end{aligned}
\end{equation}
where $\mathbf{r}_i$ is the $i$th member of the \textit{real} dataset and $N$ is the population size.
Then the GANs are trained for several iterations.
At each iteration, the discriminator is updated using three different types of training data, i.e., the \textit{real} samples, the \textit{fake} samples, and the samples generated by the generator.
The loss function for training the discriminator is given as follows:
\begin{equation}\label{eq:d1}
\begin{split}
&\max\limits_{D} V(D) = \mathbb{E}_{\mathbf{r}\in{P_{\mathbf{r}}}}[log(D(\mathbf{r}))] + \\
&\mathbb{E}_{\mathbf{f}  \in {P_{\mathbf{f}}}}[log(1-D(\mathbf{f}))]+ \mathbb{E}_{\mathbf{z}  \in {P_{\mathbf{z}}}}[log(1-D(G(\mathbf{z})))],
\end{split}
\end{equation}
where $D(\mathbf{r})$, $D(\mathbf{f})$, and $D(G(\mathbf{z}))$ denote the outputs of the discriminator with the \textit{real} sample, the \textit{fake} sample, and the sample generated by the generator being the inputs, respectively.
Specifically, both the \textit{real} and \textit{fake} samples in Algorithm 4 are fully used.
In each training iteration, we randomly divide all the data points in $\mathbf{X}$ (including both \textit{real} and \textit{fake} ones, i.e.,  $\mathbf{X}$$=$$P_{\mathbf{r}}\bigcup$$P_{\mathbf{f}}$) into several batches of size $m$.
The input of the generator is vector $\mathbf{z}$ sampled from a multivariate normal distribution.
Finally, the generator is updated according to (\ref{eq:gen2}) using the samples generated by itself.
Note that we have greedily used the imbalanced training set, aiming to enhance the convergence of GMOEA by pushing the target distribution away from the $\textit{fake}$ distribution.
In other words, we prefer a model with higher accuracy in distinguishing the fake samples, such that there is a clear margin between the target distribution and the fake one.

\begin{algorithm}[!htbp]
\caption{Model Training}
\label{al:GAN}
\begin{algorithmic}[1]
\REQUIRE
$\mathbf{X}$ (given data), $m$ (batch size).
\STATE $\boldsymbol{\mu}\leftarrow mean{(P_{\mathbf{r}})}$ /*Mean vector of $P_{\mathbf{r}}$ with $P_{\mathbf{r}}\in \mathbf{X}$*/
 \STATE $\boldsymbol {\Sigma}\leftarrow cov(P_{\mathbf{r}})$ /*Covariance matrix of $P_{\mathbf{r}}$*/\\
\FOR{total number of training iterations}
    \STATE $\mathbf{X}'\leftarrow \mathbf{X}$
    \FOR{$i\leftarrow 1:|\mathbf{X}|/m$}
        \STATE $\mathbf{T}\leftarrow$ Randomly sample $m$ data points from $\mathbf{X}'$
        \STATE $\mathbf{X'}\leftarrow \mathbf{X'}\backslash \mathbf{T}$
        \STATE $\mathbf{Z}\leftarrow$ Sample $m$ data points from multivariate normal distribution $\mathcal{N}(\boldsymbol{\mu},\boldsymbol {\Sigma})$
    	\STATE Update the discriminator according to (\ref{eq:d1}) by using $\mathbf{T}$ and $\mathbf{Z}$
    	\STATE  $\mathbf{Z}\leftarrow$ Sample $m$ data points from multivariate normal distribution $\mathcal{N}(\boldsymbol{\mu},\boldsymbol {\Sigma})$
        \STATE Update the generator according to (\ref{eq:gen2}) by using $\mathbf{Z}$
    \ENDFOR
\ENDFOR
\end{algorithmic}
\end{algorithm}

The detailed procedure of the model training in GMOEA is given in Algorithm \ref{al:GAN}.
Here, we use the multivariate normal Gaussian distribution~\cite{balakrishnan2014continuous}, which is specified by its mean vector and covariance matrix, to generate training data.
The mean vector represents the location where samples are most likely to be generated, and the covariance indicates the level to which two variables are correlated.
This modification is inspired by the generative method in variational auto-encoder (VAE)~\cite{VAE}, which aims to generate data that approximates the given distribution.
More importantly, this modification will potentially reduce the amount of data required for training the generator, since the distributions of $P_{\mathbf{z}}$ and $G(\mathbf{z})$ are similar.

\subsection{Offspring Reproduction}

In this work, we adopt a hybrid reproduction strategy for offspring generation in GMOEA, which aims at balancing the exploitation and exploration of the proposed algorithm.
The general idea of the proposed reproduction strategy is simple and efficient.
At each generation, $N$ offspring solutions will be generated either by the GAN model or the genetic operators (i.e., crossover and mutation) with equal probability.
{Since there is a risk of mode collapse in training a GAN model~\cite{arjovsky2017wasserstein}, the trained model may generate some poor solutions.
To remedy this issue, we propose to mix the candidate solutions generated by both the GAN model and genetic operators as training data.}

To generate a candidate solution using the GANs, we first calculate the mean vector $\boldsymbol{\mu}$ and covariance matrix $\boldsymbol{\Sigma}$ of the \textit{real} samples according to (\ref{eq:mean}).
Then, a $D$-dimensional vector $\mathbf {x}$ is sampled with each element being independently sampled from a continuous uniform distribution $U(0,1)$.
Afterwards, a $D$-dimensional vector $\mathbf{y}$ satisfying the multivariate normal distribution is generated according to the following probability density function:
\begin{equation}\label{eq:PDF}
\mathbf{y} =\frac{\exp{\left(-{\frac {1}{2}}(\mathbf {x} - \boldsymbol{\mu})^{\mathrm {T} }{\boldsymbol{\Sigma}}^{-1}(\mathbf {x} - \boldsymbol{\mu}) \right)}}{\sqrt{(2\pi )^{D}|\boldsymbol{\Sigma}|}},
\end{equation}
where $D$ denotes the dimensionality of the decision space.
Finally, the output of the generator, $G(\mathbf{y})$, is restricted according to the lower and upper boundaries (i.e., $\mathbf{l}$ and $\mathbf{u}$) of the decision space as follows:
$$\mathbf{x'}=G(\mathbf{y})(\mathbf{u}-\mathbf{l})+\mathbf{l},$$
where $\mathbf{x'}$ is the candidate solution generated by the GANs.

\section{Experimental Study}\label{sec:result}

To empirically examine the performance of the proposed GMOEA, we mainly conduct three different experiments to examine the properties of our proposed GMOEA.
Among these experiments, six representative MOEAs are compared, namely, NSGA-II~\cite{NSGA-II}, MOEA/D-DE~\cite{MOEADDE}, MOEA/D-CMA~\cite{MOEADCMA}, IM-MOEA\cite{IM-MOEA}, GDE3 \cite{GDE3}, and SPEA2~\cite{SPEA2}.
NSGA-II and SPEA2 are selected as they both adopt crossover and mutation operators for offspring generation.
MOEA/D-DE and GDE3 are selected as they both adopt the differential evolution operator.
MOEA/D-CMA is chosen as it is a representative MBEA, which uses the covariance matrix adaptation evolution strategy for multiobjective optimization.
Besides, IM-MOEA is selected as it is an MBEA using the inverse models to generate offspring solutions for multiobjective optimization.
The three experiments are summarized as follows:
\begin{itemize}
  \item The effectiveness of our proposed training method is examined according to the qualities of the offspring solutions generated by the GANs which are trained by different methods.
  \item The general performance of our proposed GMOEA is compared with the six algorithms on ten IMF problems with up to 200 decision variables.
  \item The effectiveness of our proposed GAN operator and the hybrid strategy is examined in comparison with the genetic operators on seven IMF problems.
\end{itemize}

In the remainder of this section, we first present a brief introduction to the experimental settings of all the compared algorithms.
Then the test problems and performance indicators are described.
Afterwards, each algorithm is run for 20 times on each test problem independently.
Then the Wilcoxon rank sum test ~\cite{haynes2013wilcoxon} is used to compare the results obtained by the proposed GMOEA and the compared algorithms at a significance level of 0.05.
Symbols `$+$', `$-$', and `$\approx$' indicate the compared algorithm is significantly better than, significantly worse than, and statistically tied by GMOEA, respectively.

\subsection{Experimental settings}

For fair comparisons, we adopt the recommended parameter settings for the compared algorithms that have achieved the best performance as reported in the literature.
The six compared algorithms are implemented in PlatEMO using Matlab~\cite{PlatEMO}, and our proposed GMOEA is implemented in Pytorch using Python 3.6.
All the algorithms are run on a PC with Intel Core i9 3.3 GHz processor, 32 GB of RAM, and 1070Ti GPU.

\textit{1) Reproduction Operators.}
In this work, the simulated binary crossover (SBX)~\cite{review-book} and the polynomial mutation (PM)~\cite{PM} are adopted for offspring generation in NSGA-II and SPEA2.
The distribution index of crossover is set to $n_c$$=$20 and the distribution index of mutation is set to $n_m$$=$20, as recommended in~\cite{review-book}.
The crossover probability $p_c$ is set to 1.0 and the mutation probability $p_m$ is set to $1/D$, where $D$ is the number of decision variables.
In MOEA/D-DE, MOEA/D-CMA, and GDE3, the differential evolution (DE) operator~\cite{DE} and PM are used for offspring generation.
Meanwhile, the control parameters are set to $CR$$=$1, $F$$=$0.5, $p_m$$=$$1/D$, and $\eta$$=$20 as recommended in~\cite{MOEADDE}.

\textit{2) Population Size.}
The population size is set to 100 for test instances with two objectives and 105 for test instances with three objectives.

\textit{(3)  Specific Parameter Settings in Each Algorithm.}
In MOEA/D-DE, the neighborhood size is set to 20, the probability of choosing parents locally is set to 0.9, and the maximum number of candidate solutions replaced by each offspring solution is set to 2.
In MOEA/D-CMA, the number of groups is set to 5.
As for IM-MOEA, the number of reference vectors is set to 10 and the size of random groups is set to 3.

In our proposed GMOEA, the training parameter settings of the GANs are fixed, where the batch size is set to 32, the learning rates for our discriminator and generator are 0.0001 and 0.0004 respectively, the total number of iterations is set to 200, and the Adam optimizer~\cite{Adam} with $\beta_{1}$$=$0.5, $\beta_{2}$$=$0.999 is used to train our GAN.
Note that the specified model in GMOEA is suitable for the benchmark investigated in this work, and its structure can be revised accordingly to fit different problems.

\textit{(4) Termination Condition.}
The total number of FEs is adopted as the termination condition for all the test instances.
The number of FEs is set to 5000 for test problems with 30 decision variables, 10000 for problems with 50 decision variables, 15000 for problems with 100 decision variables, and 30000 for problems with 200 decision variables.

\subsection{Test Problems and Performance Indicators}

In this work, we adapt ten problems selected from \cite{IM-MOEA}, termed IMF1 to IMF10.
Among these test problems, the number of objectives is three in IMF4, IMF8 and two in the rest ones.

We adopt two different performance indicators to assess the qualities of the obtained results.
The first one is IGD~\cite{IGD}, which can assess both the convergence and distribution of the obtained solution set.
Suppose that $P^*$ is a set of relatively evenly distributed reference points \cite{RPgeneration} in the PF and $\Omega$ is the set of the obtained non-dominated solutions.
The IGD can be mathematically defined as follows.
 \begin{equation}\label{eq:IGD}
\text{IGD}(P^*,\Omega)=\frac{\sum_{\mathbf{x}\in P^*}dis(\mathbf{x},\Omega)}{|P^*|},
 \end{equation}
where $dis(\mathbf{x},\Omega)$ is the minimum Euclidean distance between $\mathbf{x}$ and points in $\Omega$, and $|P^*|$ denotes the number of elements in $P^*$.
The set of reference points required for calculating IGD values are relatively evenly selected from the PF of each test problem, and a set size closest to 10000 is used in this paper.

The second performance indicator is the hypervolume (HV) indicator \cite{HV}.
Generally, hypervolume is favored because it captures in a single scalar both the closeness of the solutions to the optimal set and the spread of the solutions across objective space.
Given a solution set $\Omega$, the HV value of $\Omega$ is defined as the area covered by $\Omega$ with respect to a set of predefined reference points $P^*$ in the objective space:
\begin{equation}\label{eq:HV}
\text{HV}(\Omega,P^*)=\lambda (H(\Omega,P^*)),
\end{equation}
where
\begin{equation}
H(\Omega,P^*)=\{z\in Z|\exists x\in P,\exists r\in P^*:f(x)\leq z \leq r\} \nonumber,
\end{equation}
and $\lambda$ is the Lebesgue measure with
\begin{equation}
\lambda (H(\Omega,P^*))=\int_{\mathbb{P^*}^n}1_{H(\Omega,P^*)}(z)dz\nonumber,
\end{equation}
where $1_{H(\Omega,P^*)}$ is the characteristic function of $H(\Omega,P^*)$.

Note that, a smaller value of IGD will indicate better performance of the algorithm; in contrast, a greater value of HV will indicate better performance of the algorithm.

\subsection{Effectiveness of the Model Training Method}

To verify the effectiveness of our proposed model training method in GMOEA, we compare the offspring solutions generated by our modified GANs (where the data augmentation via multivariate Gaussian model is adopted) and the original GANs during the optimization of  IMF4 and IMF7.
We select IMF4 since its PS is complicated, and this problem is difficult for existing MOEAs to maintain diversity.
IMF7 with 200 decision variables is tested to examine the effectiveness of our proposed training method in solving MOPs with high-dimensional decision variables.
The numbers of FEs for these two problems are set to 5000 and 30000, respectively.
Besides, each test instance is tested for 10 independent runs to obtain the statistic results.
In each independent run, we sample the offspring solutions every 10 iterations for IMF4 and every 50 iterations for IMF7.

\begin{figure}[!htbp]
\centerline{\includegraphics[width=\linewidth]{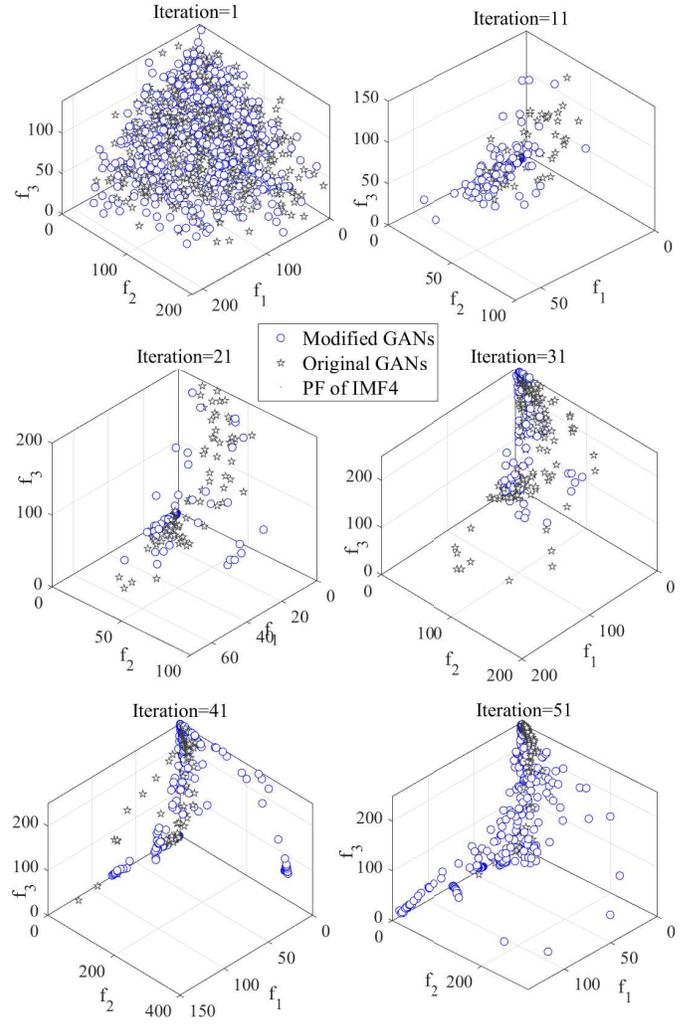}}
	\caption{The offsprings generated by the original GANs and our modified GANs at different iterations of the evolution on IMF4 with 30 decision variables.}\label{fig:compare2}
\end{figure}

\begin{figure}[!htbp]
\centerline{\includegraphics[width=\linewidth]{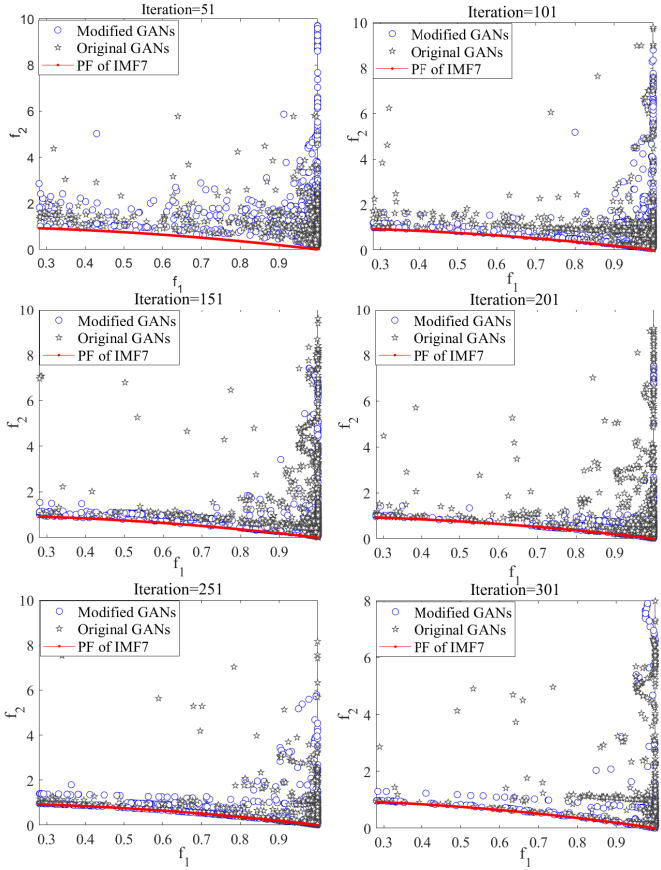}}
	\caption{The offsprings generated by the original GANs and our modified GANs at different iterations of the evolution on IMF7 with 200 decision variables.}\label{fig:compare3}
\end{figure}

Fig. \ref{fig:compare2} presents the offspring solutions obtained on tri-objective IMF4.
It can be observed that the original GANs tend to generate offspring solutions in a smaller region of the objective space (e.g., near the top center in Fig. \ref{fig:compare2}).
By contrast, our modified GANs have generated a set of widely spread offspring solutions with better convergence in most iterations.
Fig. \ref{fig:compare3} presents the offspring solutions obtained on IMF7 with 200 decision variables.
It can be observed that our modified GANs have generated a set of better-converged and spreading offspring solutions; by contrast, the original GANs have generated offspring solutions mostly in the left corner.

It can be concluded from the three comparisons that our proposed training method is effective in diversity maintenance and convergence enhancement, even on MOPs with complicated PSs and up to 200 decision variables.

\begin{figure}[!htbp]
\begin{minipage}[t]{\linewidth}
\centering
  \includegraphics[width=0.75\linewidth]{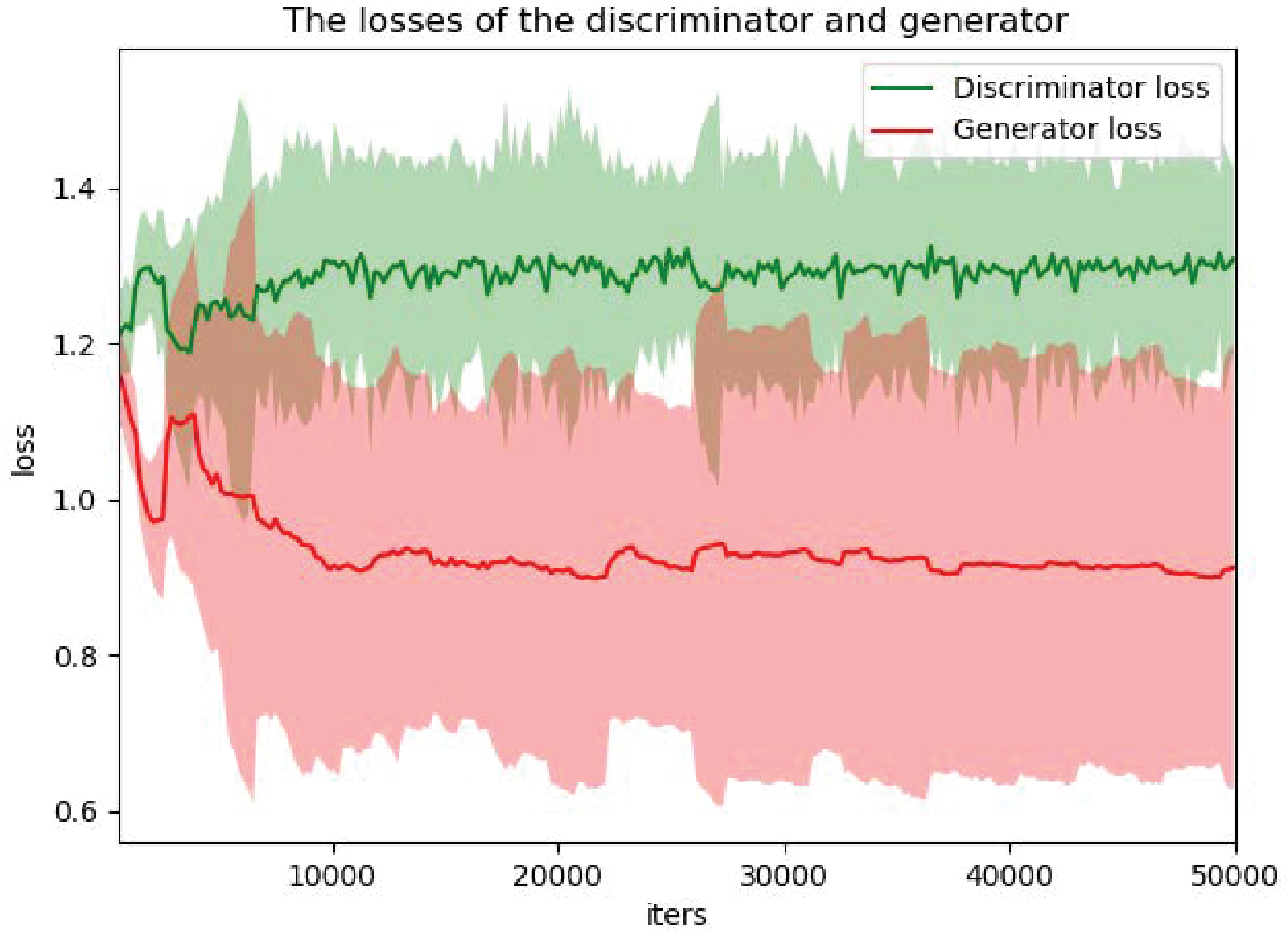}
  \includegraphics[width=0.75\linewidth]{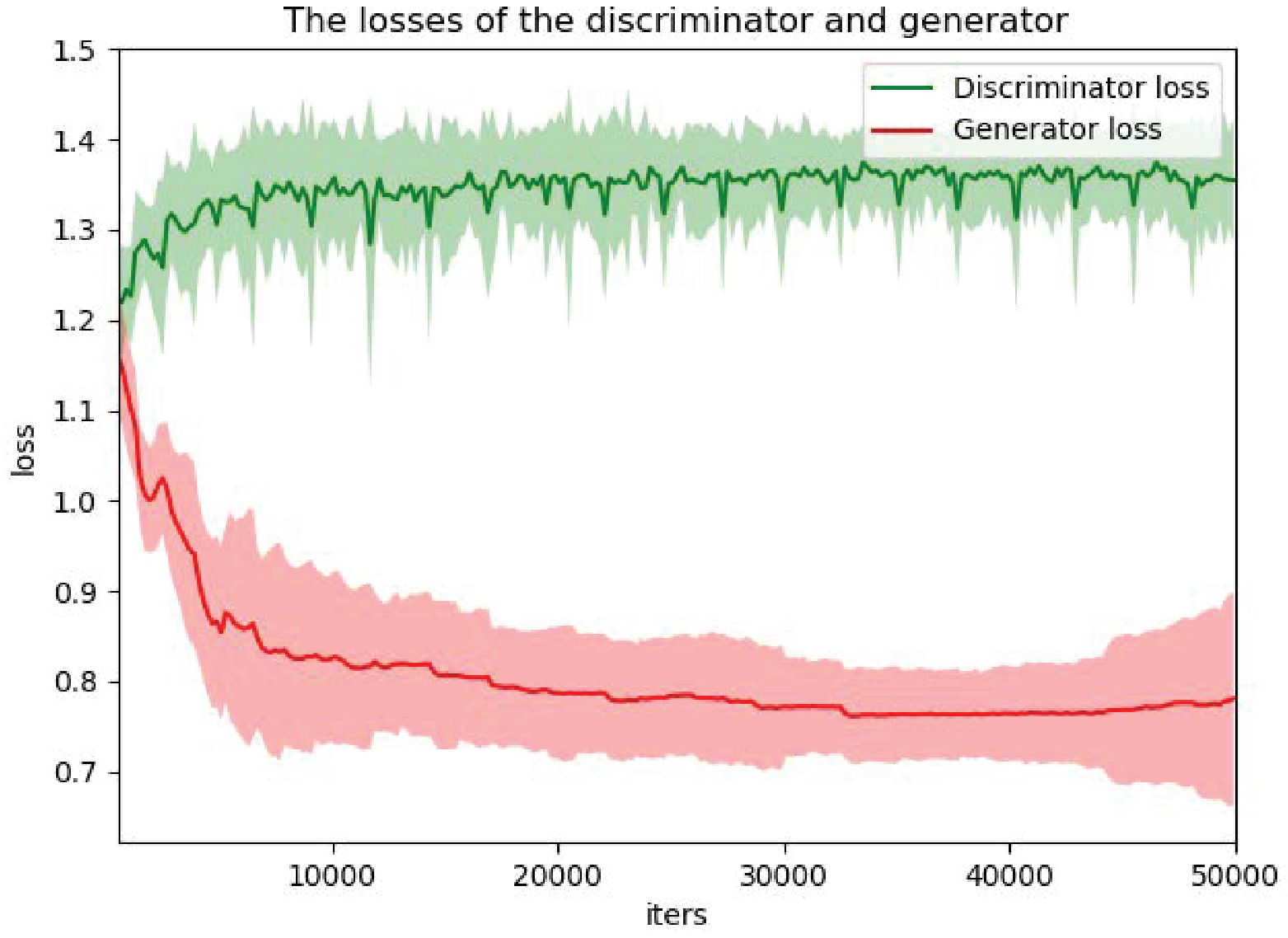}
  \end{minipage}
\caption{The trajectories of generator and discriminator's training losses of the original GAN (with multivariate Gaussian model disabled) and our modified GAN  during the evolution, respectively.}
\label{fig:traject}
    \end{figure}

Furthermore, we display the trajectories of generator and discriminator's training losses during the evolution in Fig.~\ref{fig:traject}, where GMOEA is adopted to optimize IMF1 with 30 decision variables.
In this figure, the horizontal denotes the epoch number from the first generation to the last generation of the evolution, where each epoch is averaged over 20 independent runs.
It can be observed that the training loss of each discriminator rises while the training loss of each generator drops; nevertheless, the generator in our modified GAN trends to have a lower and more stable training loss than that of the original GAN.
It can be attributed to the fact that the generator in our modified GAN generates more realistic samples that the discriminator cannot distinguish, and thus the generator is powerful in generating promising samples.
This is consistent with the design principle of offspring generators (i.e. generating promising candidate solutions) in EAs.

\subsection{General Performance}

The statistical results of the IGD and HV values achieved by the seven compared MOEAs on IMF1 to IMF10 are summarized in Table \ref{tab:IMMOEA-IGD} and Table \ref{tab:IMMOEA-HV}, respectively.
Our proposed GMOEA has performed the best on these ten problems, followed by IM-MOEA, NSGA-II, and MOEA/D-CMA.
It can be concluded from these two tables that GMOEA shows an overall better performance compared with the model-free MOEAs, i.e., NSGA-II, MOEA/D-DE, GDE3, and SPEA2, on IMF problems.
Meanwhile, GMOEA has shown a competitive performance compared with MOEA/D-CMA and IM-MOEA on these IMF problems.
\begin{table*}[!htbp]
\centering
\caption{The IGD Results Obtained by NSGA-II, MOEA/D-DE, MOEA/D-CMA, IM-MOEA, GDE3, SPEA2, and GMOEA on 40 IMF Test Instances. The Best Result in Each Row is Highlighted.}
\scalebox{0.85}{
\begin{tabular}{ccccccccc}
	\toprule
	\multirow{1}{*}{Problem}&\multirow{1}{*}{Dim}&NSGA-II&MOEA/D-DE&MOEA/D-CMA&IM-MOEA&GDE3&SPEA2&GMOEA\\
	\midrule \multirow{4}{*}{IMF1}&\multirow{1}{*}{30}&2.75e-1(3.56e-2)$+$&5.12e-1(8.51e-2)$-$&2.92e-1(4.07e-2)$+$&\hl{1.17e-1(2.86e-2)$+$}&9.92e-1(2.87e-1)$-$&2.89e-1(4.73e-2)$+$&4.46e-1(3.86e-2)\\
&\multirow{1}{*}{50}&3.13e-1(3.67e-2)$+$&5.43e-1(8.84e-2)$-$&2.26e-1(2.74e-2)$+$&\hl{1.24e-1(3.46e-2)$+$}&1.10e+0(2.43e-1)$-$&3.25e-1(3.52e-2)$+$&4.67e-1(4.44e-2)\\
&\multirow{1}{*}{100}&3.53e-1(3.20e-2)$+$&1.06e+0(1.62e-1)$-$&3.76e-1(4.08e-2)$+$&\hl{2.29e-1(3.52e-2)$+$}&2.08e+0(3.01e-1)$-$&3.85e-1(3.25e-2)$+$&4.87e-1(5.10e-2)\\
&\multirow{1}{*}{200}&3.85e-1(2.40e-2)$+$&1.29e+0(1.42e-1)$-$&4.06e-1(3.43e-2)$+$&\hl{2.61e-1(3.82e-2)$+$}&2.57e+0(2.23e-1)$-$&4.31e-1(2.63e-2)$+$&5.44e-1(5.43e-2)\\
\hline
\multirow{4}{*}{IMF2}&\multirow{1}{*}{30}&4.69e-1(5.60e-2)$+$&7.50e-1(1.67e-1)$-$&4.52e-1(7.56e-2)$+$&\hl{2.15e-1(7.97e-2)$+$}&2.01e+0(6.60e-1)$-$&4.72e-1(4.76e-2)$+$&6.10e-1(1.14e-6)\\
&\multirow{1}{*}{50}&4.78e-1(2.95e-2)$+$&7.17e-1(1.66e-1)$-$&3.28e-1(3.44e-2)$+$&\hl{2.84e-1(9.53e-2)$+$}&1.92e+0(4.60e-1)$-$&4.78e-1(2.97e-2)$+$&6.10e-1(1.14e-6)\\
&\multirow{1}{*}{100}&5.29e-1(3.08e-2)$+$&1.66e+0(3.47e-1)$-$&5.29e-1(5.96e-2)$+$&\hl{3.96e-1(5.81e-2)$+$}&3.42e+0(4.70e-1)$-$&5.67e-1(3.80e-2)$+$&6.10e-1(1.14e-6)\\
&\multirow{1}{*}{200}&5.75e-1(3.95e-2)$+$&2.28e+0(2.47e-1)$-$&5.93e-1(6.10e-2)$\approx$&\hl{4.11e-1(3.16e-2)$+$}&4.28e+0(3.25e-1)$-$&6.48e-1(5.17e-2)$+$&6.81e-1(2.29e-1)\\
\hline
\multirow{4}{*}{IMF3}&\multirow{1}{*}{30}&3.02e-1(9.01e-2)$-$&6.73e-1(2.58e-1)$-$&4.61e-1(5.61e-2)$-$&1.58e-1(2.97e-2)$-$&2.26e+0(5.06e-1)$-$&3.25e-1(9.57e-2)$-$&\hl{1.00e-2(1.78e-8)}\\
&\multirow{1}{*}{50}&1.74e-1(4.43e-2)$-$&7.59e-1(2.02e-1)$-$&3.89e-1(2.07e-2)$-$&1.29e-1(3.33e-2)$-$&2.96e+0(5.62e-1)$-$&2.12e-1(5.22e-2)$-$&\hl{1.00e-2(1.78e-8)}\\
&\multirow{1}{*}{100}&3.26e-1(6.12e-2)$-$&2.03e+0(2.83e-1)$-$&5.41e-1(4.80e-2)$-$&2.42e-1(4.59e-2)$-$&5.78e+0(8.25e-1)$-$&3.57e-1(8.06e-2)$-$&\hl{1.00e-2(1.78e-8)}\\
&\multirow{1}{*}{200}&3.76e-1(6.04e-2)$-$&2.74e+0(2.63e-1)$-$&5.72e-1(5.99e-2)$-$&2.37e-1(2.83e-2)$-$&7.25e+0(6.55e-1)$-$&4.11e-1(5.90e-2)$-$&\hl{7.60e-2(1.60e-1)}\\
\hline
\multirow{4}{*}{IMF4}&\multirow{1}{*}{30}&1.17e+0(3.48e-1)$-$&2.75e+0(8.05e-1)$-$&1.43e+0(2.53e-1)$-$&2.18e+0(4.68e-1)$-$&7.19e+0(2.53e+0)$-$&1.21e+0(3.11e-1)$-$&\hl{5.21e-1(1.23e-2)}\\
&\multirow{1}{*}{50}&1.54e+0(4.63e-1)$-$&3.78e+0(1.08e+0)$-$&1.44e+0(2.36e-1)$-$&2.96e+0(4.95e-1)$-$&1.85e+1(5.25e+0)$-$&1.47e+0(4.31e-1)$-$&\hl{5.37e-1(5.71e-3)}\\
&\multirow{1}{*}{100}&6.62e+0(1.56e+0)$-$&1.91e+1(3.76e+0)$-$&4.71e+0(8.46e-1)$-$&1.33e+1(2.19e+0)$-$&8.23e+1(1.59e+1)$-$&5.84e+0(9.71e-1)$-$&\hl{6.57e-1(5.23e-1)}\\
&\multirow{1}{*}{200}&1.95e+1(2.25e+0)$-$&4.65e+1(5.99e+0)$-$&9.62e+0(1.79e+0)$-$&3.21e+1(5.33e+0)$-$&2.14e+2(1.82e+1)$-$&1.52e+1(1.88e+0)$-$&\hl{1.01e+0(2.08e+0)}\\
\hline
\multirow{4}{*}{IMF5}&\multirow{1}{*}{30}&9.90e-2(1.02e-2)$-$&1.39e-1(8.13e-3)$-$&1.40e-1(1.32e-2)$-$&\hl{7.55e-2(8.87e-3)$\approx$}&1.10e-1(2.21e-2)$-$&9.70e-2(1.08e-2)$-$&7.55e-2(1.10e-2)\\
&\multirow{1}{*}{50}&1.08e-1(1.28e-2)$-$&1.35e-1(1.15e-2)$-$&1.33e-1(1.42e-2)$-$&\hl{6.80e-2(6.16e-3)$+$}&1.29e-1(1.17e-2)$-$&1.09e-1(1.04e-2)$-$&8.15e-2(1.35e-2)\\
&\multirow{1}{*}{100}&1.37e-1(8.75e-3)$\approx$&1.68e-1(7.68e-3)$-$&1.62e-1(8.13e-3)$-$&\hl{1.02e-1(6.16e-3)$\approx$}&1.68e-1(8.94e-3)$-$&1.43e-1(7.33e-3)$-$&1.20e-1(3.20e-2)\\
&\multirow{1}{*}{200}&1.60e-1(9.45e-3)$-$&1.85e-1(5.13e-3)$-$&1.66e-1(9.99e-3)$-$&1.13e-1(7.33e-3)$\approx$&1.88e-1(6.96e-3)$-$&1.75e-1(1.73e-2)$-$&\hl{1.11e-1(1.80e-2)}\\
\hline
\multirow{4}{*}{IMF6}&\multirow{1}{*}{30}&1.77e-1(2.32e-2)$-$&1.93e-1(1.87e-2)$-$&1.92e-1(1.39e-2)$-$&\hl{1.01e-1(1.17e-2)$+$}&1.61e-1(4.18e-2)$-$&1.80e-1(1.84e-2)$-$&1.17e-1(1.53e-2)\\
&\multirow{1}{*}{50}&1.92e-1(2.19e-2)$-$&1.94e-1(2.50e-2)$-$&1.80e-1(2.78e-2)$-$&\hl{9.70e-2(8.01e-3)$+$}&1.96e-1(2.21e-2)$-$&2.02e-1(1.93e-2)$-$&1.25e-1(1.15e-2)\\
&\multirow{1}{*}{100}&2.70e-1(2.83e-2)$-$&2.37e-1(8.01e-3)$-$&2.23e-1(1.49e-2)$-$&\hl{1.41e-1(6.86e-3)$\approx$}&2.59e-1(1.36e-2)$-$&2.79e-1(2.52e-2)$-$&1.77e-1(7.11e-2)\\
&\multirow{1}{*}{200}&3.19e-1(2.48e-2)$-$&2.59e-1(5.53e-3)$-$&2.45e-1(1.23e-2)$-$&\hl{1.54e-1(6.81e-3)$+$}&2.80e-1(7.95e-3)$-$&3.32e-1(3.62e-2)$-$&1.90e-1(3.50e-2)\\
\hline
\multirow{4}{*}{IMF7}&\multirow{1}{*}{30}&1.79e-1(1.79e-2)$-$&2.83e-1(1.16e-2)$-$&2.87e-1(5.87e-3)$-$&2.45e-1(7.61e-3)$-$&3.00e-1(1.03e-2)$-$&1.98e-1(2.59e-2)$-$&\hl{6.40e-2(3.03e-2)}\\
&\multirow{1}{*}{50}&1.58e-1(1.94e-2)$-$&2.83e-1(6.57e-3)$-$&2.84e-1(5.03e-3)$-$&2.32e-1(1.28e-2)$-$&2.94e-1(8.26e-3)$-$&1.64e-1(2.09e-2)$-$&\hl{5.10e-2(5.48e-2)}\\
&\multirow{1}{*}{100}&2.03e-1(2.20e-2)$-$&2.91e-1(2.24e-3)$-$&2.93e-1(4.89e-3)$-$&2.50e-1(6.49e-3)$-$&3.05e-1(6.07e-3)$-$&2.09e-1(1.59e-2)$-$&\hl{1.65e-2(1.46e-2)}\\
&\multirow{1}{*}{200}&2.39e-1(2.02e-2)$-$&2.94e-1(5.10e-3)$-$&2.95e-1(5.13e-3)$-$&2.53e-1(8.65e-3)$-$&3.08e-1(5.23e-3)$-$&2.42e-1(1.98e-2)$-$&\hl{7.25e-2(8.58e-2)}\\
\hline
\multirow{4}{*}{IMF8}&\multirow{1}{*}{30}&7.37e-1(1.18e-1)$-$&6.44e-1(3.60e-2)$-$&6.12e-1(1.05e-1)$-$&5.59e-1(4.83e-2)$-$&6.92e-1(1.81e-1)$-$&7.44e-1(1.24e-1)$-$&\hl{3.41e-1(1.90e-2)}\\
&\multirow{1}{*}{50}&9.80e-1(1.20e-1)$-$&6.71e-1(2.85e-2)$-$&6.81e-1(4.24e-2)$-$&6.55e-1(4.56e-2)$-$&9.33e-1(7.64e-2)$-$&1.00e+0(1.48e-1)$-$&\hl{3.58e-1(1.15e-2)}\\
&\multirow{1}{*}{100}&1.74e+0(1.62e-1)$-$&7.35e-1(5.38e-2)$-$&7.74e-1(3.13e-2)$-$&1.28e+0(7.10e-2)$-$&1.72e+0(2.64e-1)$-$&2.43e+0(2.16e-1)$-$&\hl{4.85e-1(8.46e-2)}\\
&\multirow{1}{*}{200}&4.00e+0(6.32e-1)$-$&\hl{8.55e-1(1.10e-1)$\approx$}&8.88e-1(3.26e-2)$\approx$&2.28e+0(2.16e-1)$-$&3.40e+0(4.49e-1)$-$&5.96e+0(4.13e-1)$-$&1.31e+0(1.55e+0)\\
\hline
\multirow{4}{*}{IMF9}&\multirow{1}{*}{30}&1.10e-1(1.49e-2)$-$&2.91e-1(5.01e-2)$-$&3.28e-1(5.37e-2)$-$&2.09e-1(2.20e-2)$-$&2.50e-1(3.63e-2)$-$&1.17e-1(1.38e-2)$-$&\hl{7.30e-2(2.64e-2)}\\
&\multirow{1}{*}{50}&1.07e-1(1.92e-2)$-$&2.91e-1(4.21e-2)$-$&3.70e-1(4.42e-2)$-$&1.78e-1(2.61e-2)$-$&2.87e-1(4.50e-2)$-$&1.10e-1(1.08e-2)$-$&\hl{8.75e-2(2.69e-2)}\\
&\multirow{1}{*}{100}&1.46e-1(9.40e-3)$-$&4.44e-1(4.83e-2)$-$&4.80e-1(3.34e-2)$-$&2.89e-1(2.61e-2)$-$&3.80e-1(4.26e-2)$-$&1.48e-1(8.75e-3)$-$&\hl{1.16e-1(3.27e-2)}\\
&\multirow{1}{*}{200}&1.73e-1(8.01e-3)$-$&5.50e-1(2.03e-2)$-$&5.26e-1(3.35e-2)$-$&2.95e-1(2.91e-2)$-$&4.96e-1(2.54e-2)$-$&1.71e-1(1.04e-2)$-$&\hl{1.41e-1(5.95e-2)}\\
\hline
\multirow{4}{*}{IMF10}&\multirow{1}{*}{30}&6.13e+1(1.76e+1)$-$&6.99e+1(1.20e+1)$-$&7.06e+1(8.84e+0)$-$&\hl{3.05e+1(9.21e+0)$+$}&1.09e+2(1.89e+1)$-$&5.07e+1(1.04e+1)$-$&3.94e+1(4.22e+0)\\
&\multirow{1}{*}{50}&1.06e+2(2.03e+1)$-$&1.18e+2(2.36e+1)$-$&1.46e+2(2.62e+1)$-$&\hl{5.26e+1(1.18e+1)$+$}&2.19e+2(2.22e+1)$-$&9.44e+1(1.51e+1)$-$&6.25e+1(4.97e+0)\\
&\multirow{1}{*}{100}&3.03e+2(3.21e+1)$-$&3.23e+2(3.90e+1)$-$&4.12e+2(4.89e+1)$-$&1.33e+2(3.49e+1)$\approx$&5.11e+2(5.13e+1)$-$&2.89e+2(5.16e+1)$-$&\hl{1.23e+2(2.48e+1)}\\
&\multirow{1}{*}{200}&6.54e+2(8.88e+1)$-$&7.19e+2(5.55e+1)$-$&9.48e+2(5.80e+1)$-$&\hl{3.29e+2(8.25e+1)$\approx$}&1.18e+3(9.33e+1)$-$&7.29e+2(9.98e+1)$-$&4.00e+2(2.29e+2)\\
\hline
\multicolumn{2}{c}{$+/-/\approx$}&8/31/1&0/39/1&7/31/2&14/20/6&0/40/0&8/32/0&\\
\bottomrule
\end{tabular}
}
\begin{tablenotes}
\footnotesize
\item["\dag"] '$+$', '$-$' and '$\approx$' indicate that the result is significantly better, significantly worse and statistically similar to that obtained by GMOEA, respectively.
\end{tablenotes}
\label{tab:IMMOEA-IGD}
\end{table*}

\begin{table*}[!htbp]
\centering
\caption{The HV Results Obtained by NSGA-II, MOEA/D-DE, MOEA/D-CMA, IM-MOEA, GDE3, SPEA2, and GMOEA on 40 IMF Test Instances. The Best Result in Each Row is Highlighted.}
\scalebox{0.85}{
\begin{tabular}{ccccccccc}
	\toprule
	\multirow{1}{*}{Problem}&\multirow{1}{*}{Dim}&NSGA-II&MOEA/D-DE&MOEA/D-CMA&IM-MOEA&GDE3&SPEA2&GMOEA\\
	\midrule \multirow{4}{*}{IMF1}&\multirow{1}{*}{30}&5.43e-1(3.16e-2)$+$&1.85e-1(6.65e-2)$-$&4.00e-1(4.98e-2)$-$&\hl{7.18e-1(1.99e-2)$+$}&2.45e-2(2.87e-2)$-$&5.19e-1(4.64e-2)$\approx$&5.08e-1(2.55e-2)\\
&\multirow{1}{*}{50}&5.50e-1(3.43e-2)$+$&1.69e-1(6.34e-2)$-$&4.99e-1(4.09e-2)$\approx$&\hl{7.29e-1(1.90e-2)$+$}&1.50e-2(4.01e-2)$-$&5.45e-1(1.93e-2)$+$&4.87e-1(3.26e-2)\\
&\multirow{1}{*}{100}&4.80e-1(3.68e-2)$\approx$&6.00e-3(1.76e-2)$-$&3.11e-1(3.75e-2)$-$&\hl{6.29e-1(2.16e-2)$+$}&0.00e+0(0.00e+0)$-$&4.46e-1(4.22e-2)$\approx$&4.65e-1(3.50e-2)\\
&\multirow{1}{*}{200}&4.33e-1(2.75e-2)$\approx$&0.00e+0(0.00e+0)$-$&2.77e-1(2.98e-2)$-$&\hl{6.06e-1(2.39e-2)$+$}&0.00e+0(0.00e+0)$-$&3.80e-1(3.59e-2)$-$&4.07e-1(8.18e-2)\\
\hline
\multirow{4}{*}{IMF2}&\multirow{1}{*}{30}&4.65e-2(3.13e-2)$-$&1.70e-2(3.69e-2)$-$&1.00e-1(4.24e-2)$\approx$&\hl{2.71e-1(5.94e-2)$+$}&0.00e+0(0.00e+0)$-$&4.05e-2(3.50e-2)$-$&1.10e-1(2.85e-7)\\
&\multirow{1}{*}{50}&5.90e-2(2.45e-2)$-$&2.15e-2(2.85e-2)$-$&1.92e-1(2.88e-2)$+$&\hl{2.32e-1(5.86e-2)$+$}&0.00e+0(0.00e+0)$-$&5.35e-2(2.64e-2)$-$&1.10e-1(2.85e-7)\\
&\multirow{1}{*}{100}&6.50e-3(7.45e-3)$-$&0.00e+0(0.00e+0)$-$&6.70e-2(2.75e-2)$-$&\hl{1.41e-1(2.92e-2)$+$}&0.00e+0(0.00e+0)$-$&0.00e+0(0.00e+0)$-$&1.10e-1(2.85e-7)\\
&\multirow{1}{*}{200}&0.00e+0(0.00e+0)$-$&0.00e+0(0.00e+0)$-$&4.35e-2(2.30e-2)$-$&\hl{1.36e-1(1.39e-2)$+$}&0.00e+0(0.00e+0)$-$&0.00e+0(0.00e+0)$-$&9.35e-2(4.03e-2)\\
\hline
\multirow{4}{*}{IMF3}&\multirow{1}{*}{30}&1.32e-1(5.32e-2)$-$&2.85e-2(6.27e-2)$-$&4.75e-2(1.92e-2)$-$&2.39e-1(2.88e-2)$-$&0.00e+0(0.00e+0)$-$&1.14e-1(5.15e-2)$-$&\hl{4.24e-1(5.03e-3)}\\
&\multirow{1}{*}{50}&2.34e-1(4.20e-2)$-$&7.50e-3(1.12e-2)$-$&7.60e-2(1.14e-2)$-$&2.69e-1(3.20e-2)$-$&0.00e+0(0.00e+0)$-$&1.99e-1(4.23e-2)$-$&\hl{4.28e-1(4.10e-3)}\\
&\multirow{1}{*}{100}&1.19e-1(3.75e-2)$-$&0.00e+0(0.00e+0)$-$&2.50e-2(1.10e-2)$-$&1.71e-1(3.43e-2)$-$&0.00e+0(0.00e+0)$-$&1.01e-1(4.12e-2)$-$&\hl{4.24e-1(5.03e-3)}\\
&\multirow{1}{*}{200}&9.05e-2(3.03e-2)$-$&0.00e+0(0.00e+0)$-$&1.95e-2(1.23e-2)$-$&1.73e-1(2.15e-2)$-$&0.00e+0(0.00e+0)$-$&7.45e-2(2.72e-2)$-$&\hl{3.64e-1(1.26e-1)}\\
\hline
\multirow{4}{*}{IMF4}&\multirow{1}{*}{30}&5.00e-4(2.24e-3)$-$&0.00e+0(0.00e+0)$-$&0.00e+0(0.00e+0)$-$&0.00e+0(0.00e+0)$-$&0.00e+0(0.00e+0)$-$&3.50e-3(1.18e-2)$-$&\hl{4.35e-1(2.26e-2)}\\
&\multirow{1}{*}{50}&0.00e+0(0.00e+0)$-$&0.00e+0(0.00e+0)$-$&0.00e+0(0.00e+0)$-$&0.00e+0(0.00e+0)$-$&0.00e+0(0.00e+0)$-$&0.00e+0(0.00e+0)$-$&\hl{4.54e-1(9.95e-3)}\\
&\multirow{1}{*}{100}&0.00e+0(0.00e+0)$-$&0.00e+0(0.00e+0)$-$&0.00e+0(0.00e+0)$-$&0.00e+0(0.00e+0)$-$&0.00e+0(0.00e+0)$-$&0.00e+0(0.00e+0)$-$&\hl{4.35e-1(1.02e-1)}\\
&\multirow{1}{*}{200}&0.00e+0(0.00e+0)$-$&0.00e+0(0.00e+0)$-$&0.00e+0(0.00e+0)$-$&0.00e+0(0.00e+0)$-$&0.00e+0(0.00e+0)$-$&0.00e+0(0.00e+0)$-$&\hl{4.35e-1(1.02e-1)}\\
\hline
\multirow{4}{*}{IMF5}&\multirow{1}{*}{30}&7.03e-1(1.39e-2)$-$&6.13e-1(1.14e-2)$-$&6.09e-1(1.85e-2)$-$&7.23e-1(1.42e-2)$-$&6.66e-1(4.81e-2)$-$&7.04e-1(1.23e-2)$-$&\hl{7.58e-1(1.69e-2)}\\
&\multirow{1}{*}{50}&7.07e-1(1.21e-2)$-$&6.23e-1(2.08e-2)$-$&6.23e-1(2.30e-2)$-$&7.40e-1(9.99e-3)$-$&6.28e-1(1.84e-2)$-$&7.06e-1(1.15e-2)$-$&\hl{7.53e-1(1.89e-2)}\\
&\multirow{1}{*}{100}&6.60e-1(1.05e-2)$-$&5.70e-1(1.03e-2)$-$&5.83e-1(9.10e-3)$-$&6.83e-1(1.17e-2)$-$&5.74e-1(1.23e-2)$-$&6.54e-1(9.88e-3)$-$&\hl{7.10e-1(3.50e-2)}\\
&\multirow{1}{*}{200}&6.32e-1(1.01e-2)$-$&5.46e-1(5.98e-3)$-$&5.78e-1(1.21e-2)$-$&6.68e-1(8.13e-3)$-$&5.46e-1(9.40e-3)$-$&6.14e-1(1.43e-2)$-$&\hl{7.13e-1(3.28e-2)}\\
\hline
\multirow{4}{*}{IMF6}&\multirow{1}{*}{30}&2.98e-1(2.31e-2)$-$&3.39e-1(2.07e-2)$-$&3.39e-1(1.70e-2)$-$&4.04e-1(1.35e-2)$\approx$&3.72e-1(4.76e-2)$-$&2.95e-1(1.76e-2)$-$&\hl{4.08e-1(1.48e-2)}\\
&\multirow{1}{*}{50}&2.80e-1(2.24e-2)$-$&3.35e-1(2.52e-2)$-$&3.50e-1(2.64e-2)$-$&4.03e-1(9.67e-3)$\approx$&3.34e-1(2.66e-2)$-$&2.70e-1(2.08e-2)$-$&\hl{4.08e-1(1.37e-2)}\\
&\multirow{1}{*}{100}&1.97e-1(2.80e-2)$-$&2.95e-1(1.10e-2)$-$&3.08e-1(1.61e-2)$\approx$&\hl{3.60e-1(9.45e-3)$+$}&2.67e-1(1.53e-2)$-$&1.86e-1(2.35e-2)$-$&3.16e-1(7.80e-2)\\
&\multirow{1}{*}{200}&1.57e-1(1.89e-2)$-$&2.73e-1(6.57e-3)$-$&2.83e-1(1.49e-2)$\approx$&\hl{3.47e-1(8.01e-3)$+$}&2.46e-1(8.21e-3)$-$&1.42e-1(3.27e-2)$-$&2.97e-1(3.92e-2)\\
\hline
\multirow{4}{*}{IMF7}&\multirow{1}{*}{30}&2.35e-1(1.54e-2)$-$&1.62e-1(8.34e-3)$-$&1.58e-1(5.23e-3)$-$&1.91e-1(6.41e-3)$-$&1.38e-1(8.51e-3)$-$&2.19e-1(2.16e-2)$-$&\hl{3.46e-1(4.31e-2)}\\
&\multirow{1}{*}{50}&2.54e-1(2.06e-2)$-$&1.61e-1(2.24e-3)$-$&1.61e-1(5.10e-3)$-$&2.03e-1(1.02e-2)$-$&1.45e-1(6.07e-3)$-$&2.49e-1(2.34e-2)$-$&\hl{3.69e-1(6.40e-2)}\\
&\multirow{1}{*}{100}&2.10e-1(2.22e-2)$-$&1.51e-1(3.08e-3)$-$&1.50e-1(6.49e-3)$-$&1.84e-1(6.81e-3)$-$&1.37e-1(4.70e-3)$-$&2.06e-1(1.57e-2)$-$&\hl{4.11e-1(2.24e-2)}\\
&\multirow{1}{*}{200}&1.78e-1(1.74e-2)$-$&1.49e-1(3.08e-3)$-$&1.48e-1(5.23e-3)$-$&1.80e-1(9.45e-3)$-$&1.40e-1(5.70e-17)$-$&1.77e-1(1.49e-2)$-$&\hl{3.46e-1(9.47e-2)}\\
\hline
\multirow{4}{*}{IMF8}&\multirow{1}{*}{30}&3.50e-3(8.13e-3)$-$&1.43e-1(2.07e-2)$-$&1.71e-1(1.10e-1)$-$&8.50e-2(3.07e-2)$-$&1.53e-1(1.66e-1)$-$&3.50e-3(8.13e-3)$-$&\hl{3.55e-1(6.92e-2)}\\
&\multirow{1}{*}{50}&0.00e+0(0.00e+0)$-$&1.28e-1(1.92e-2)$-$&1.20e-1(1.38e-2)$-$&2.00e-2(2.10e-2)$-$&2.50e-3(1.12e-2)$-$&0.00e+0(0.00e+0)$-$&\hl{4.66e-1(4.80e-2)}\\
&\multirow{1}{*}{100}&0.00e+0(0.00e+0)$-$&5.15e-2(4.37e-2)$-$&1.40e-2(1.23e-2)$-$&0.00e+0(0.00e+0)$-$&0.00e+0(0.00e+0)$-$&0.00e+0(0.00e+0)$-$&\hl{3.40e-1(1.21e-1)}\\
&\multirow{1}{*}{200}&0.00e+0(0.00e+0)$-$&1.05e-2(1.85e-2)$-$&0.00e+0(0.00e+0)$-$&0.00e+0(0.00e+0)$-$&0.00e+0(0.00e+0)$-$&0.00e+0(0.00e+0)$-$&\hl{2.24e-1(2.20e-1)}\\
\hline
\multirow{4}{*}{IMF9}&\multirow{1}{*}{30}&6.77e-1(2.08e-2)$-$&4.43e-1(5.79e-2)$-$&3.92e-1(6.29e-2)$-$&5.32e-1(3.54e-2)$-$&4.86e-1(4.84e-2)$-$&6.63e-1(2.03e-2)$-$&\hl{7.78e-1(3.24e-2)}\\
&\multirow{1}{*}{50}&6.87e-1(3.10e-2)$-$&4.41e-1(4.96e-2)$-$&3.46e-1(4.49e-2)$-$&5.82e-1(3.93e-2)$-$&4.38e-1(5.59e-2)$-$&6.81e-1(1.81e-2)$-$&\hl{7.60e-1(2.75e-2)}\\
&\multirow{1}{*}{100}&6.28e-1(1.28e-2)$-$&2.78e-1(4.39e-2)$-$&2.41e-1(2.70e-2)$-$&4.25e-1(3.17e-2)$-$&3.32e-1(4.29e-2)$-$&6.23e-1(1.26e-2)$-$&\hl{7.37e-1(2.23e-2)}\\
&\multirow{1}{*}{200}&5.89e-1(1.12e-2)$-$&1.88e-1(1.54e-2)$-$&2.07e-1(2.64e-2)$-$&4.20e-1(3.51e-2)$-$&2.26e-1(1.79e-2)$-$&5.90e-1(1.61e-2)$-$&\hl{7.01e-1(5.72e-2)}\\
\hline
\multirow{4}{*}{IMF10}&\multirow{1}{*}{30}&0.00e+0(0.00e+0)$\approx$&0.00e+0(0.00e+0)$\approx$&0.00e+0(0.00e+0)$\approx$&0.00e+0(0.00e+0)$\approx$&0.00e+0(0.00e+0)$\approx$&0.00e+0(0.00e+0)$\approx$&0.00e+0(0.00e+0)\\
&\multirow{1}{*}{50}&0.00e+0(0.00e+0)$\approx$&0.00e+0(0.00e+0)$\approx$&0.00e+0(0.00e+0)$\approx$&0.00e+0(0.00e+0)$\approx$&0.00e+0(0.00e+0)$\approx$&0.00e+0(0.00e+0)$\approx$&0.00e+0(0.00e+0)\\
&\multirow{1}{*}{100}&0.00e+0(0.00e+0)$\approx$&0.00e+0(0.00e+0)$\approx$&0.00e+0(0.00e+0)$\approx$&0.00e+0(0.00e+0)$\approx$&0.00e+0(0.00e+0)$\approx$&0.00e+0(0.00e+0)$\approx$&0.00e+0(0.00e+0)\\
&\multirow{1}{*}{200}&0.00e+0(0.00e+0)$\approx$&0.00e+0(0.00e+0)$\approx$&0.00e+0(0.00e+0)$\approx$&0.00e+0(0.00e+0)$\approx$&0.00e+0(0.00e+0)$\approx$&0.00e+0(0.00e+0)$\approx$&0.00e+0(0.00e+0)\\
\hline
\multicolumn{2}{c}{$+/-/\approx$}&2/32/6&0/36/4&1/31/8&10/24/6&0/36/4&1/33/6&\\
\bottomrule
\end{tabular}
}
\begin{tablenotes}
\footnotesize
\item["\dag"] '$+$', '$-$' and '$\approx$' indicate that the result is significantly better, significantly worse and statistically similar to that obtained by GMOEA, respectively.
\end{tablenotes}
\label{tab:IMMOEA-HV}
\end{table*}

\begin{figure*}[!htbp]
	\centerline{\includegraphics[width=\linewidth]{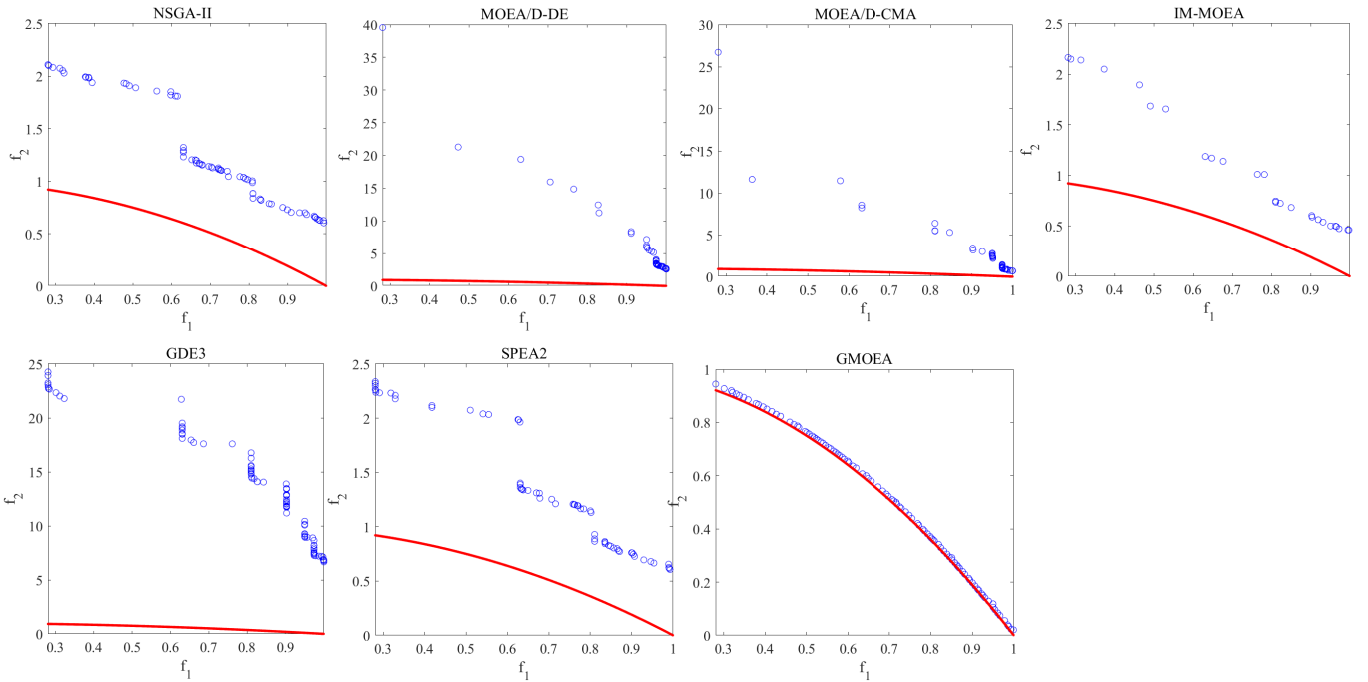}}
	\caption{The final non-dominated solutions obtained by the compared algorithms on bi-objective IMF3 with 200 decision variables in the run associated with the median IGD value.}
	\label{fig:IMF3}
\end{figure*}

\begin{figure*}[!htbp]
	\centerline{\includegraphics[width=\linewidth]{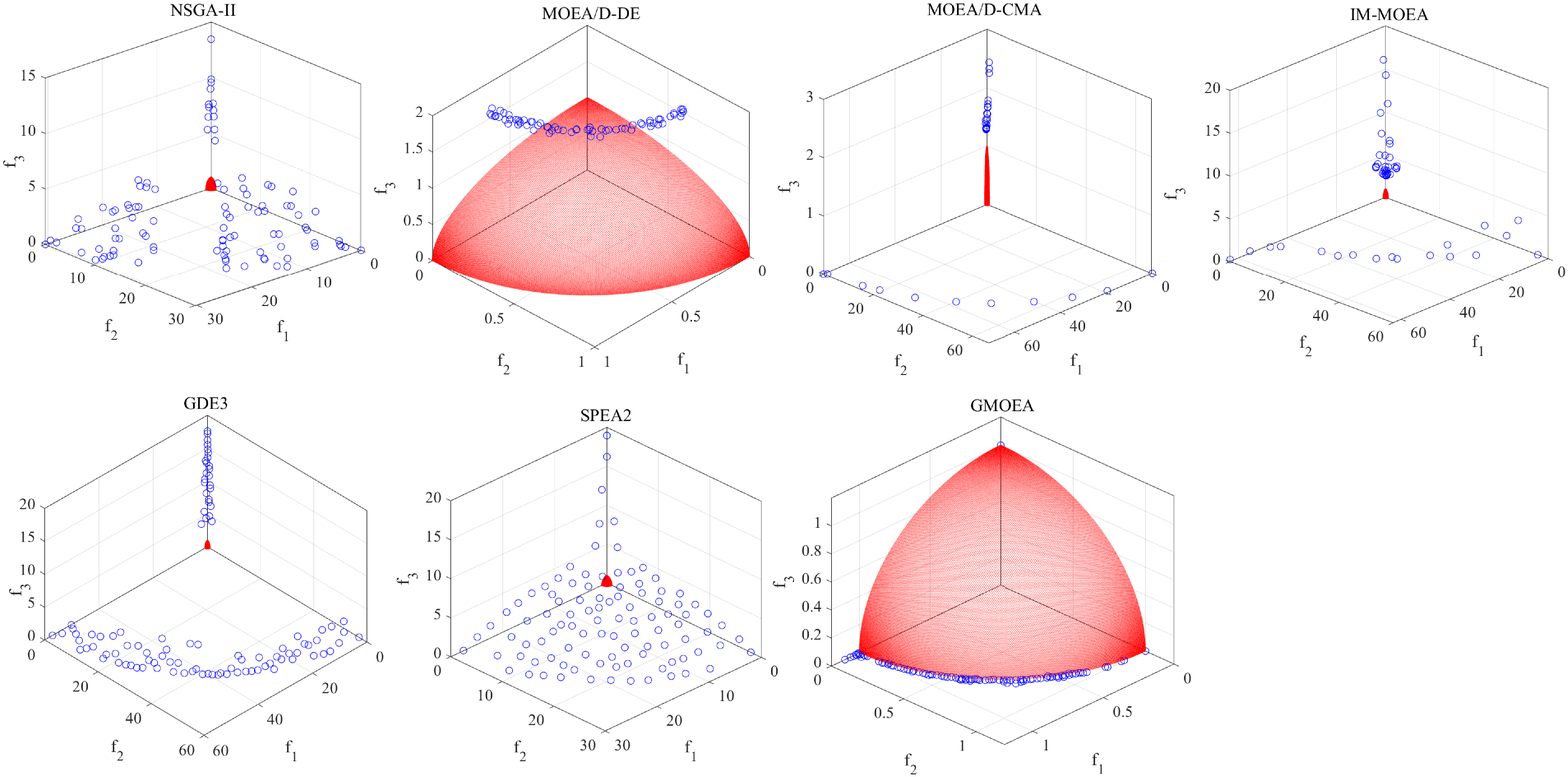}}
	\caption{The final non-dominated solutions obtained by the compared algorithms on bi-objective IMF8 with 200 decision variables in the run associated with the median IGD value.}
	\label{fig:IMF8}
\end{figure*}

The final non-dominated solutions achieved by the compared algorithms on bi-objective IMF3 and tri-objective IMF8 with 200 decision variables in the runs associated with the median IGD value are plotted in Fig. \ref{fig:IMF3} and Fig. \ref{fig:IMF8}, respectively.
It can be observed that GMOEA has achieved the best results on these problems, where the obtained non-dominated solutions are best converged.

\begin{figure*}[!htbp]
	\centering
	\subfigure{
		\begin{minipage}[t]{0.33\linewidth}
			\centering
			\includegraphics[width=\linewidth]{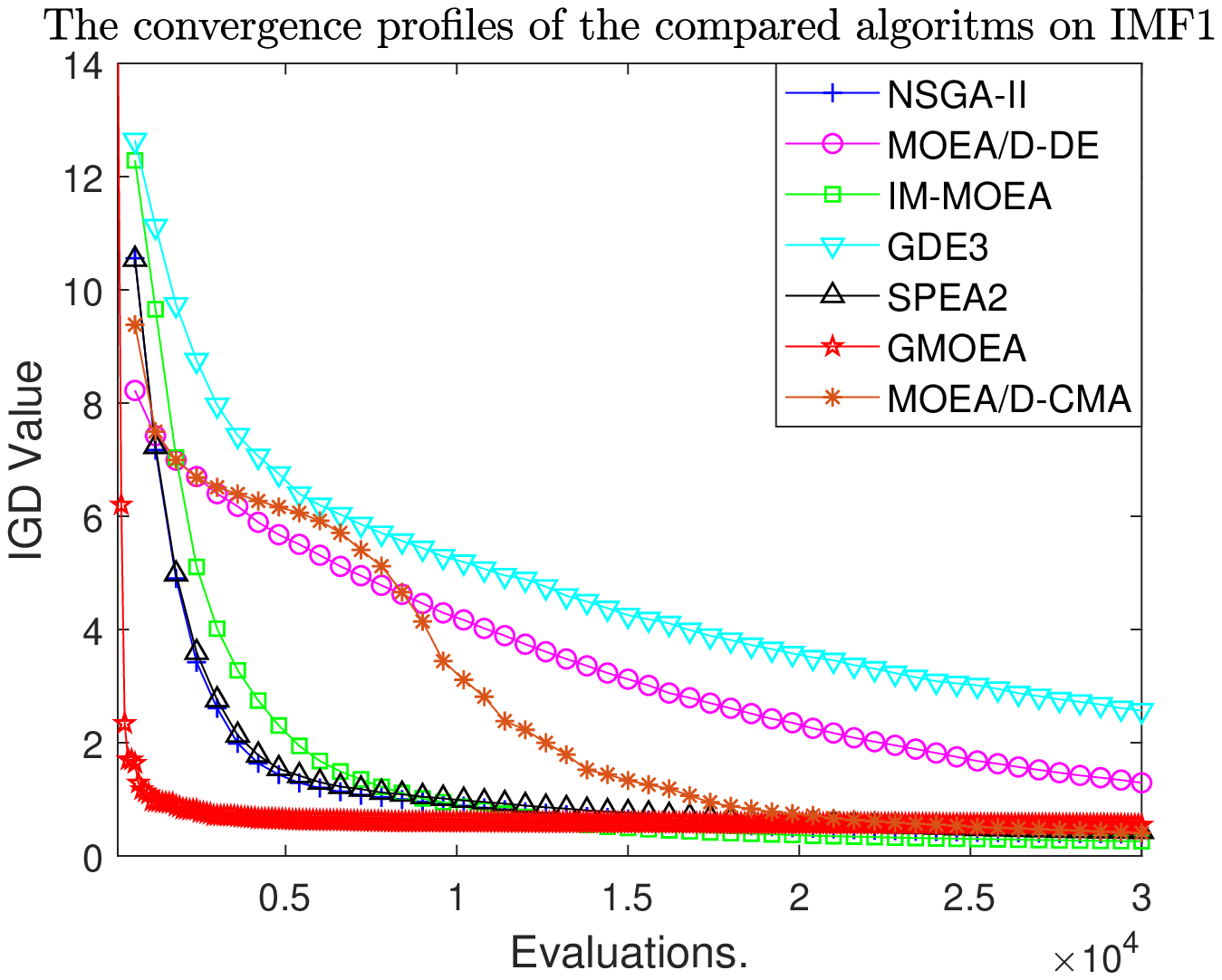}
		\end{minipage}
		\begin{minipage}[t]{0.33\linewidth}
			\centering
			\includegraphics[width=\linewidth]{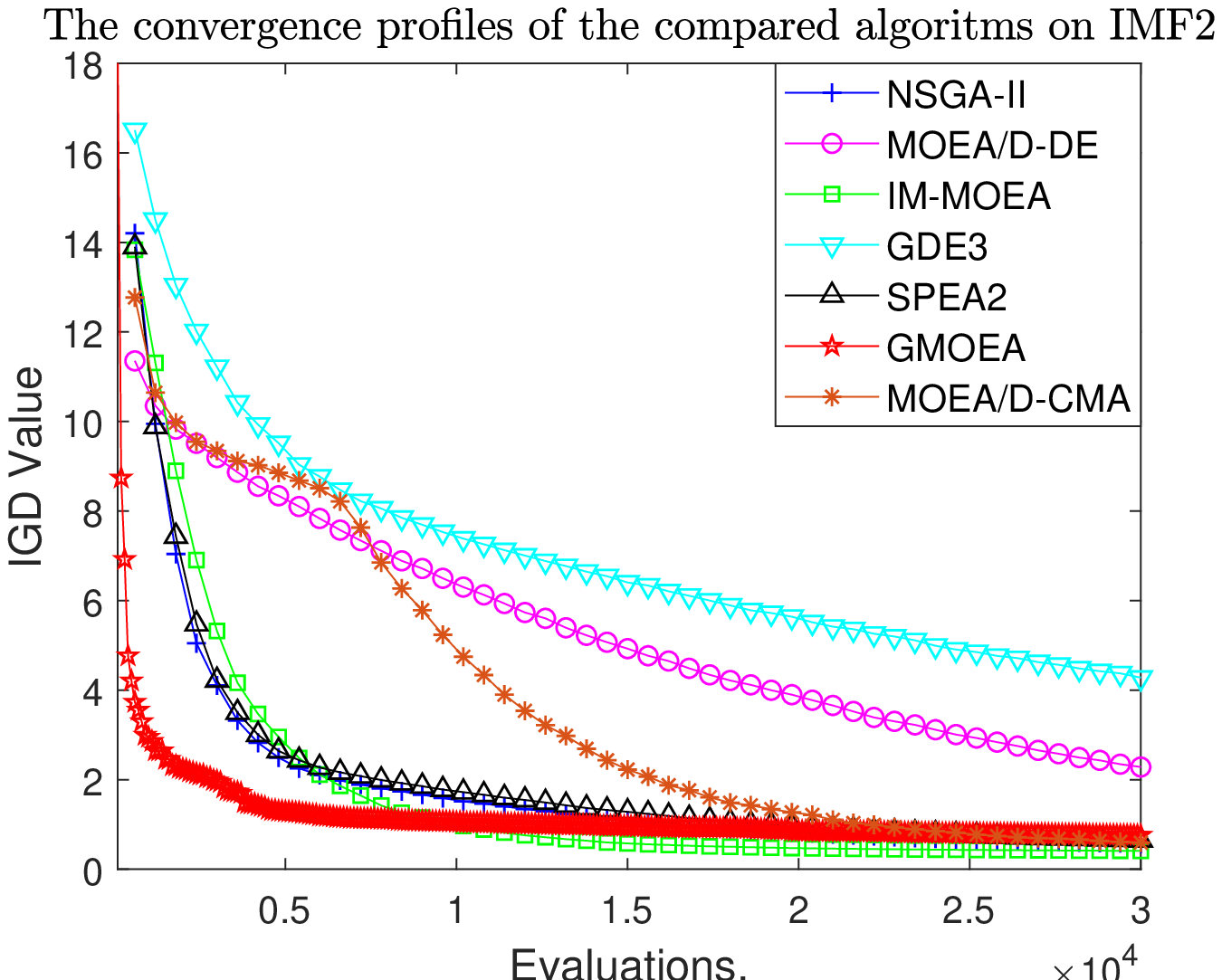}
		\end{minipage}
		\begin{minipage}[t]{0.33\linewidth}
		\centering
		\includegraphics[width=\linewidth]{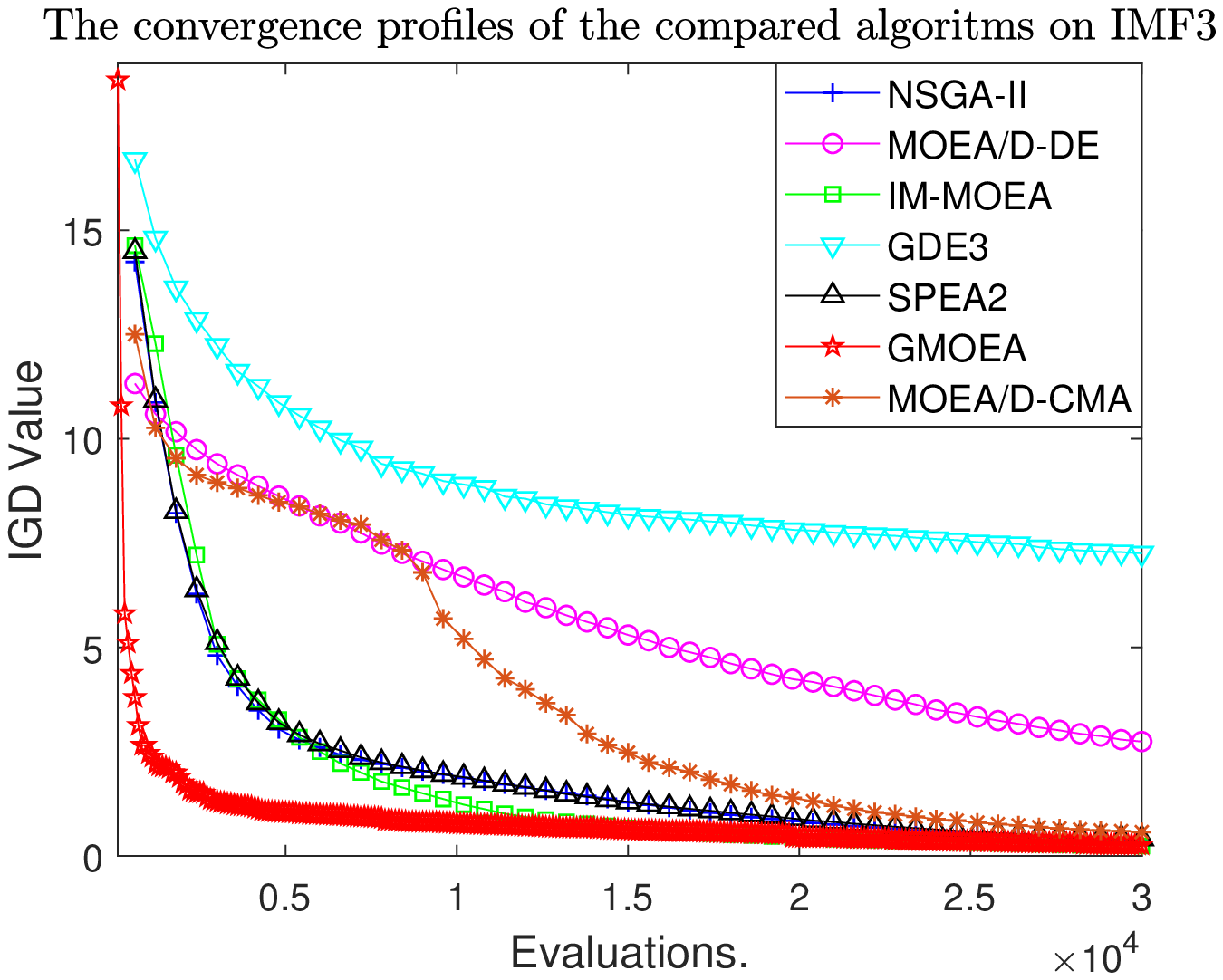}
	\end{minipage}
	}
	\subfigure{
	\begin{minipage}[t]{0.33\linewidth}
		\centering
		\includegraphics[width=\linewidth]{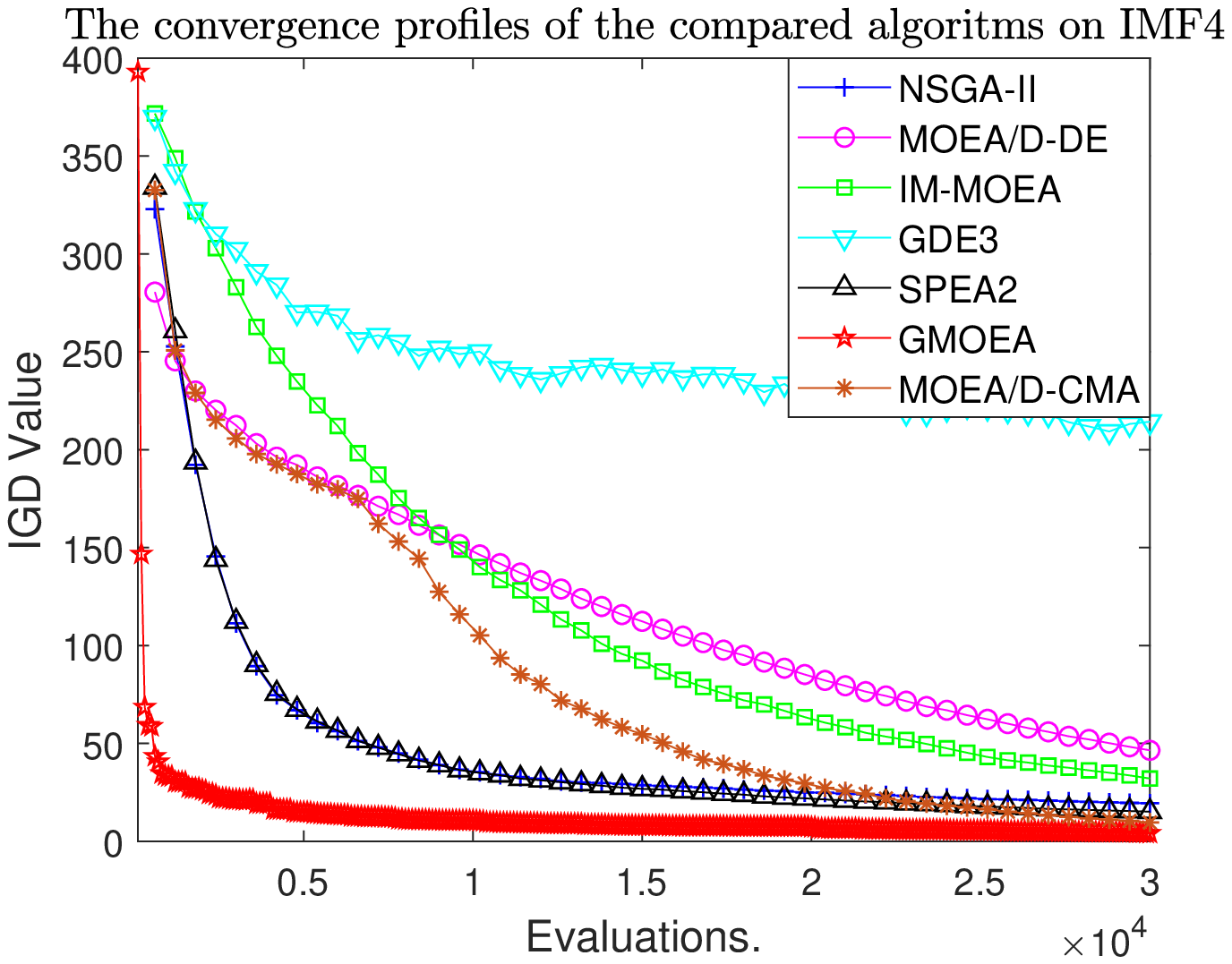}
	\end{minipage}
	\begin{minipage}[t]{0.33\linewidth}
		\centering
		\includegraphics[width=\linewidth]{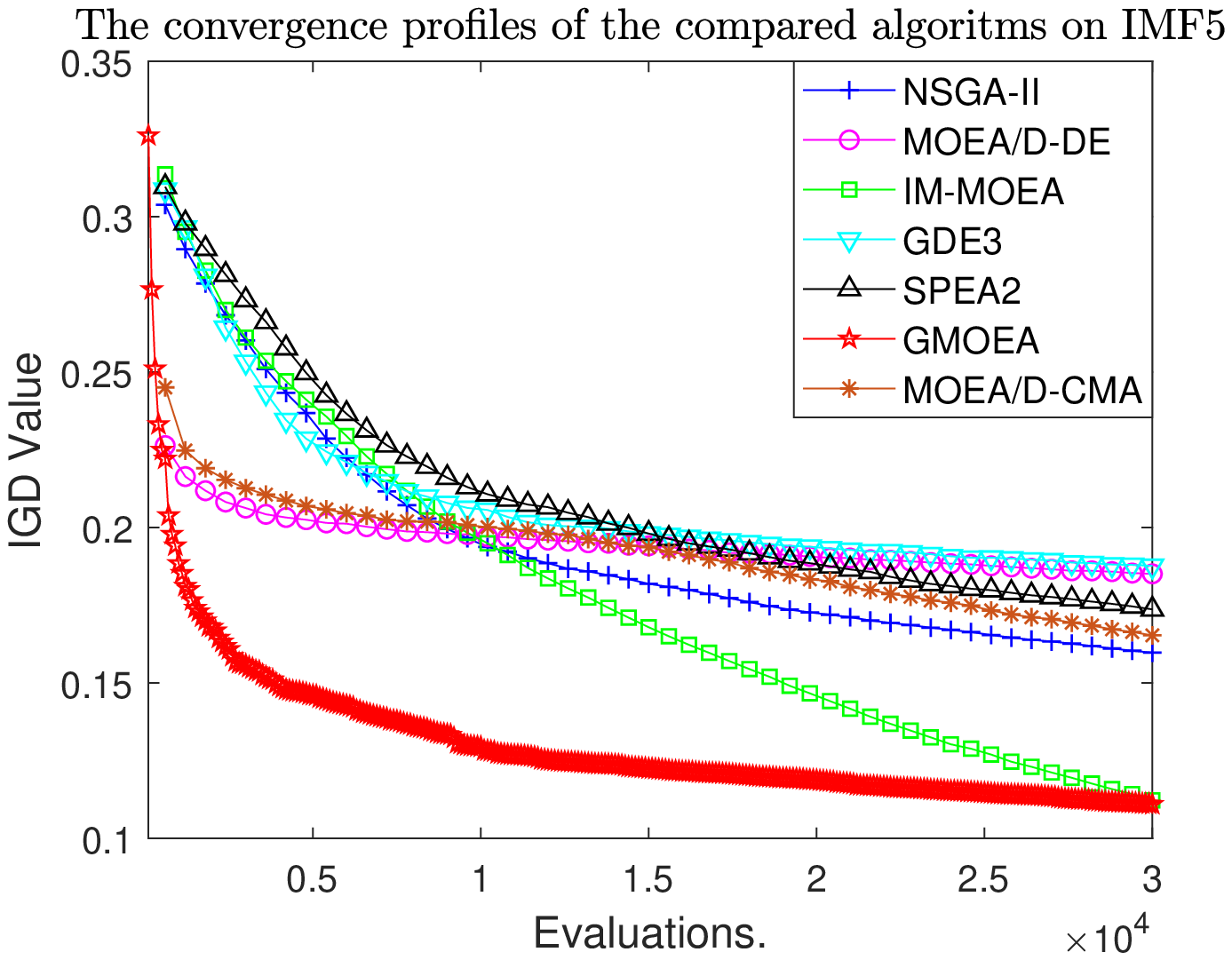}
	\end{minipage}
	\begin{minipage}[t]{0.33\linewidth}
		\centering
		\includegraphics[width=\linewidth]{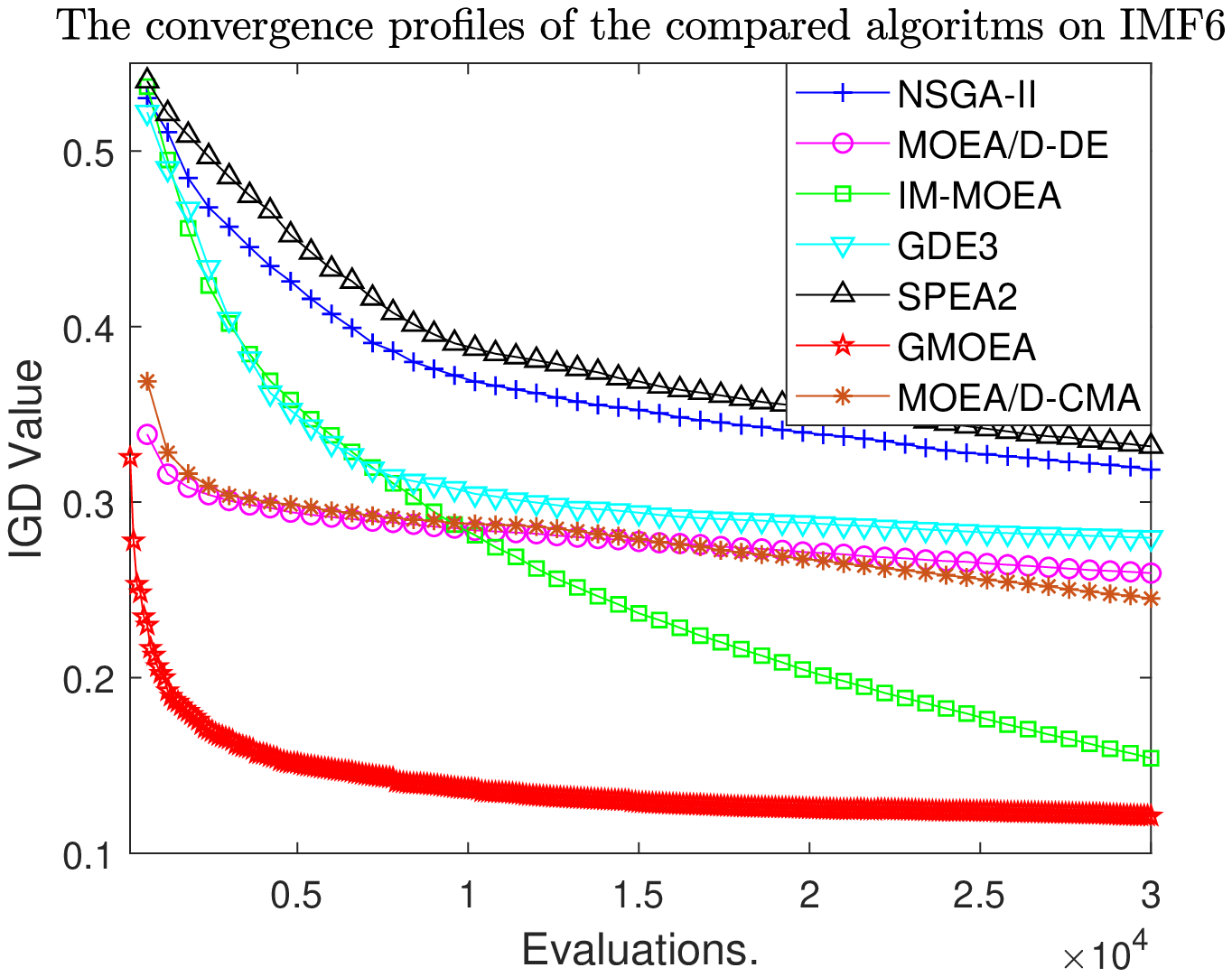}
	\end{minipage}
}
	\subfigure{
	\begin{minipage}[t]{0.33\linewidth}
		\centering
		\includegraphics[width=\linewidth]{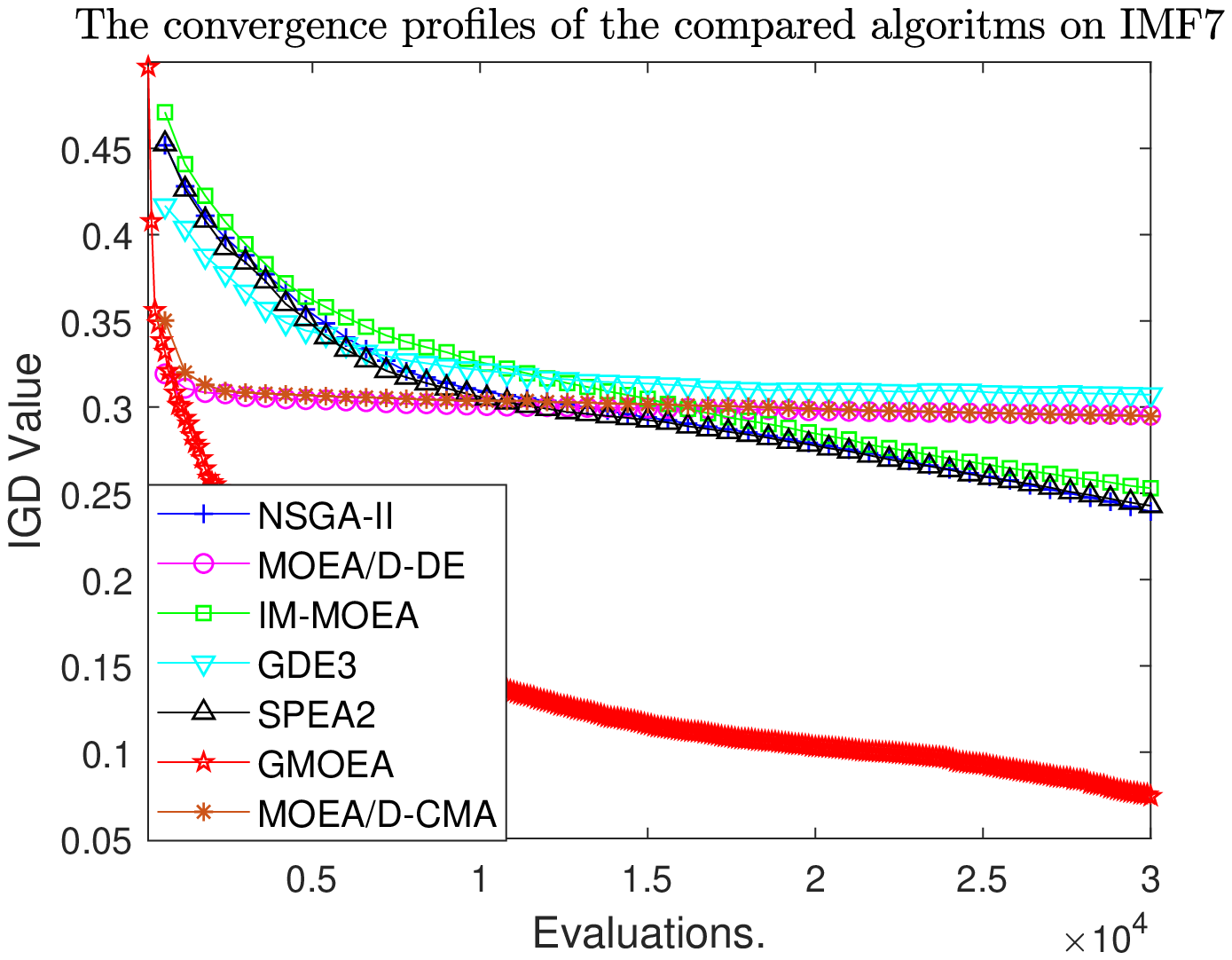}
	\end{minipage}
	\begin{minipage}[t]{0.33\linewidth}
		\centering
		\includegraphics[width=\linewidth]{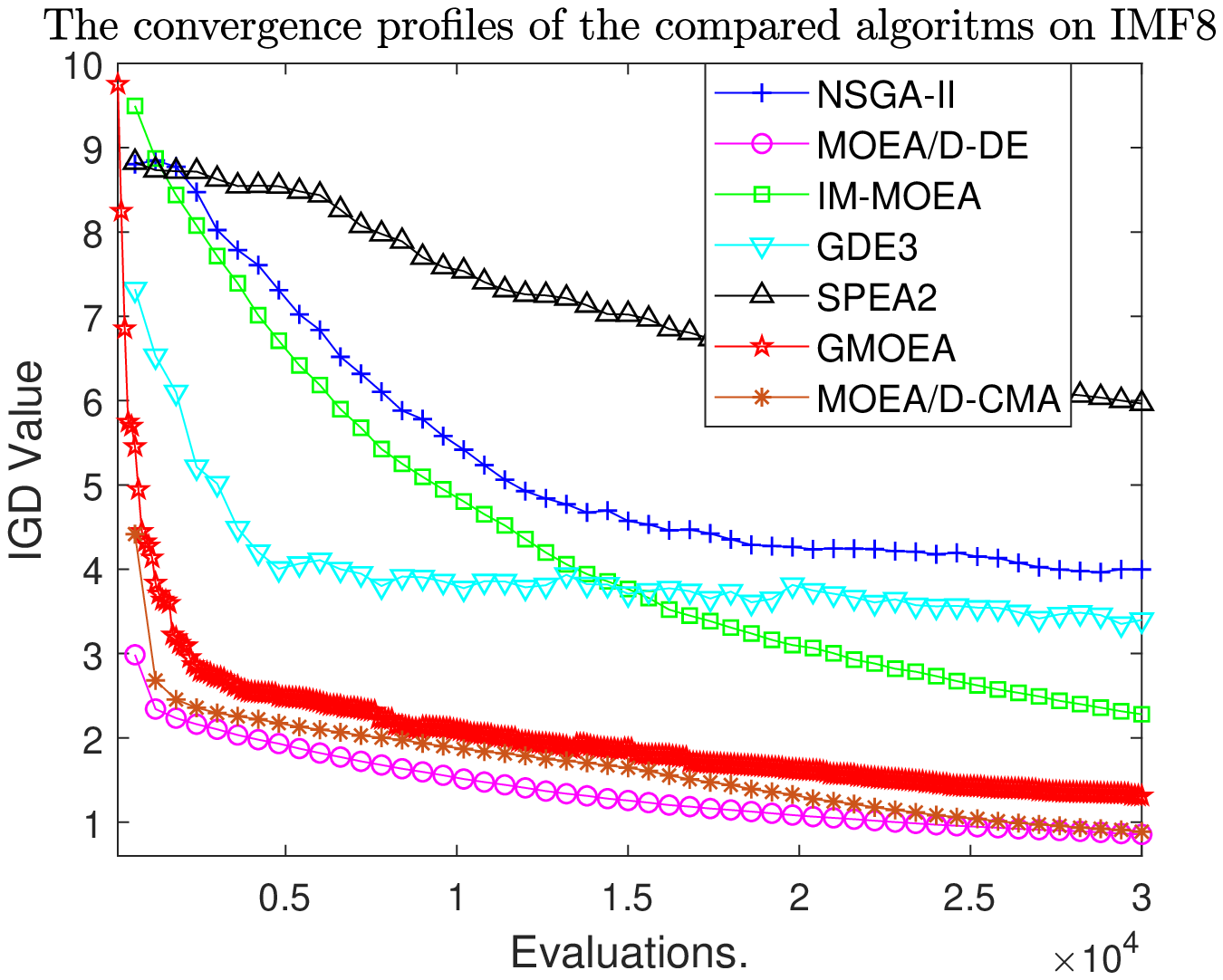}
	\end{minipage}
	\begin{minipage}[t]{0.33\linewidth}
		\centering
		\includegraphics[width=\linewidth]{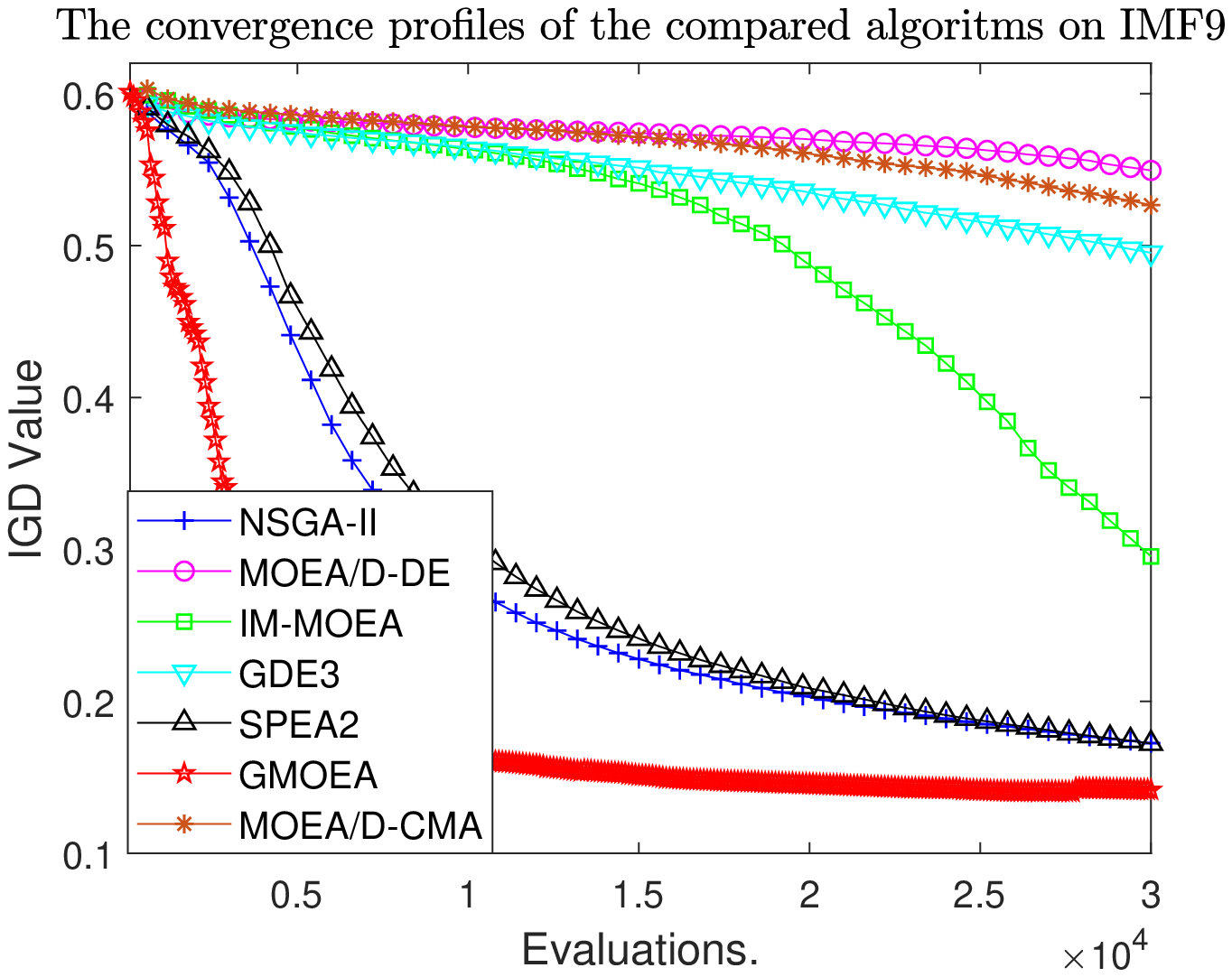}
	\end{minipage}
}
	\caption{The convergence profiles of the seven compared algorithms on IMF1 to IMF9 with 200 decision variables, respectively. }
	\label{fig:IMF_Convergence}
\end{figure*}

The convergence profiles of the seven compared algorithms on nine IMF problems with 200 decision variables are given in Fig~\ref{fig:IMF_Convergence}.
It can be observed that GMOEA converges faster than the other six compared algorithms on most problems.
The results have demonstrated the superiority of our proposed GMOEA over the six compared algorithms on MOPs with up to 200 decision variables in terms of convergence speed.

\begin{figure}[!htbp]
	\centerline{\includegraphics[width=0.9\linewidth]{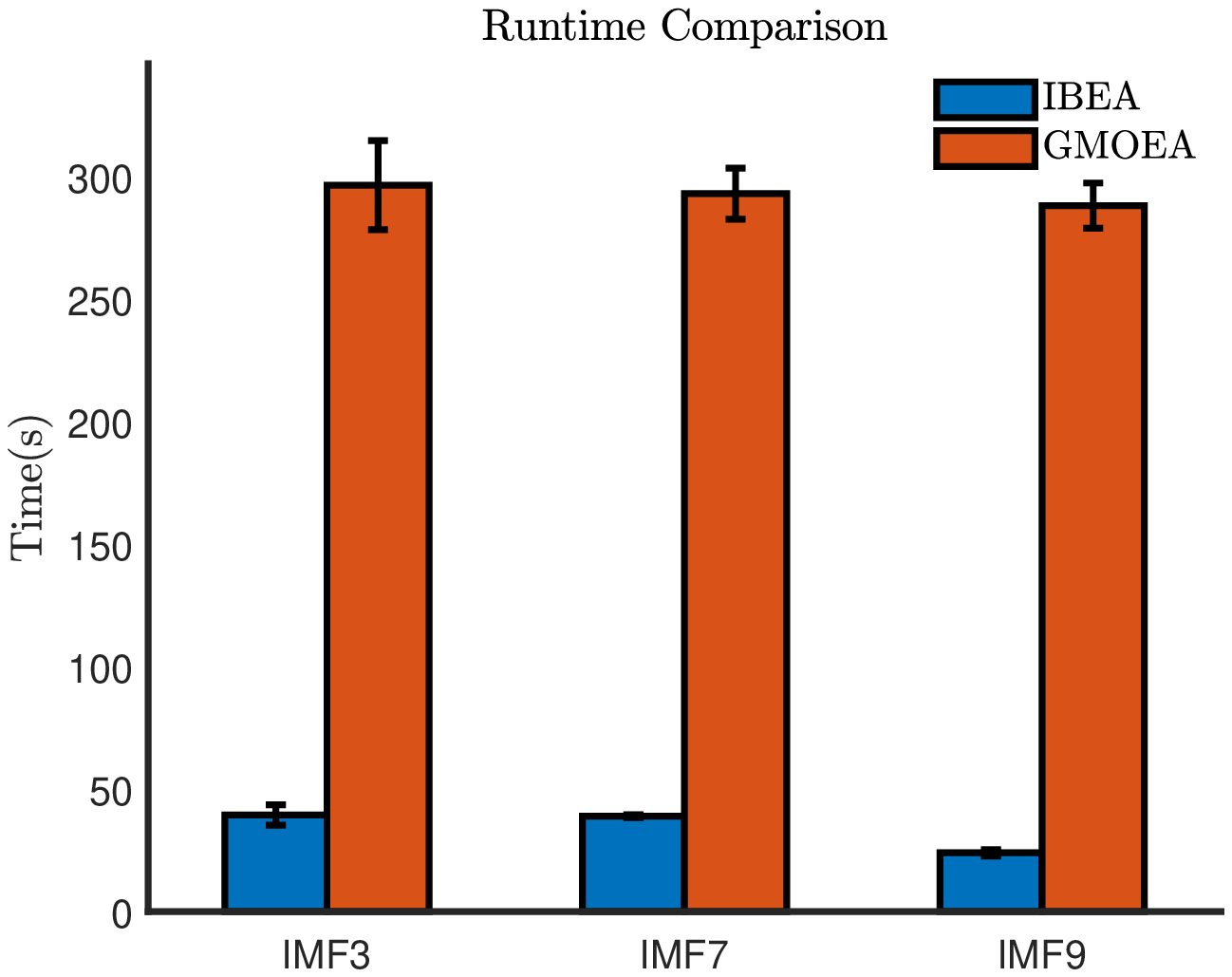}}
  \caption{The statistics of the runtime results achieved by the original IBEA and GMOEA.}
  \label{fig:runtime}
\end{figure}

Since our GMOEA is implemented in Python on Pytorch~\cite{pytorch}, while the compared ones are implemented in Matlab on PlatEMO~\cite{PlatEMO}, the runtime comparison among them could be unfair.
Nevertheless, we have conducted a comparison between GMOEA (embedded in IBEA) and the standard IBEA both in Python.
The runtime achieved by each algorithm on three IMF problems with 30 decision variables is presented in Fig.~\ref{fig:runtime}.
It can be observed that the runtime of GMOEA is about five times as much as that of IBEA, which can be further improved by using some high-performance GPU rather than the NVIDIA 1070 Ti as we did in this work.
As an offline optimizer, such a time cost is generally acceptable.

\subsection{Ablation Study}
Here, we further investigate the performance of pure genetic operators (i.e., the reproduction without GAN, termed GMOEA$*$), pure GAN operator (i.e., the reproduction without crossover or mutation, termed GMOEA$-$), and the hybrid operator (i.e., the original GMOEA) on IMF3 to IMF8 with 30, 50, 100, and 200 decision variables, respectively.

\begin{figure*}[!htbp]
	\centerline{\includegraphics[width=\linewidth]{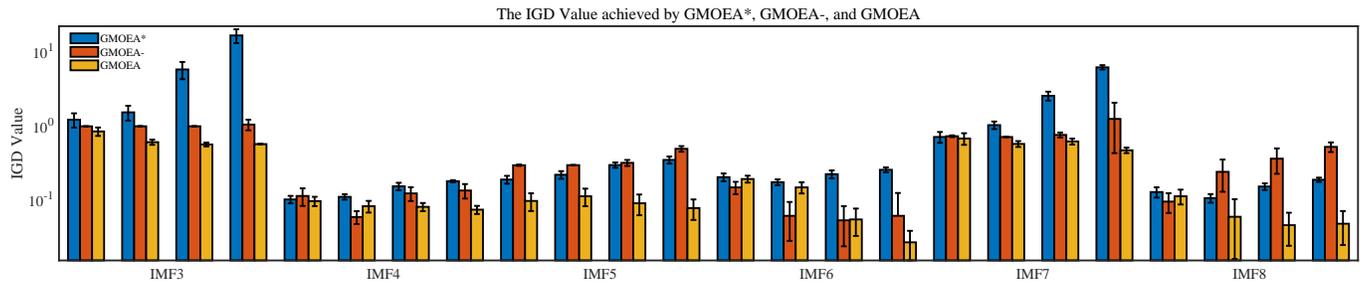}}
  \caption{The statistics of IGD results achieved by GMOEA$*$ (the reproduction with pure genetic operators), GMOEA$-$ (the reproduction with pure GAN operator), and GMOEA (the reproduction with the hybrid strategy) on seven IMF problems with a number of 30, 50, 100, and 200 decision variables, respectively.}
  \label{fig:compare}
\end{figure*}

The statistics of IGD results achieved by these three compared algorithms are given in Fig.~\ref{fig:compare}.
As indicated by the results, the pure GAN operator and the hybrid one perform significantly better than pure genetic operators on almost all the test instances, and GMOEA outperforms GMOEA$-$ on most test instances.
Hence, the proposed GAN operator coupled with the hybrid strategy is effective in handling MOPs.

\section{Conclusion}\label{sec:conclusion}

In this work, we have proposed an MOEA driven by the GANs, termed GMOEA, for solving MOPs with up to 200 decision variables.
Due to the learning and generative abilities of the GANs, GMOEA is effective in solving these problems.

The GANs in GMOEA are adopted for generating promising offspring solutions under the framework of MBEAs.
In GMOEA, we first classify candidate solutions in the current population into two different datasets, where some high-quality candidate solutions are labeled as \textit{real} samples and the rest are labeled as \textit{fake} samples.
Since the GANs mimic the distribution of target data, the distribution of \textit{real} samples should consider two issues.
The first issue is the diversity of training data, which ensures that the data could represent the general distribution of the expected solutions.
The second issue is the convergence of training data, which ensures that the generated samples could satisfy the target of minimizing all the objectives.

A novel training method is proposed in GMOEA to take full advantage of the two datasets.
During the training,  both the \textit{real} and \textit{fake} datasets, as well as the data generated by the generator,  are used to train the discriminator.
It is highlighted that the proposed training method is demonstrated to be powerful and effective.
Only a relatively small amount of training data is used for training the GANs (a total number of 100 samples for an MOP with 2 objectives and 105 samples for MOPs with 3 objectives).
Besides, we also introduce an offspring reproduction strategy to further improve the performance of our proposed GMOEA.
By hybridizing the classic stochastic reproduction and generating sampling based reproduction, the exploitation and exploration can be balanced.

To assess the performance of our proposed GMOEA, some empirical comparisons have been conducted on a set of MOPs with up to 200 decision variables.
The general performance of our proposed GMOEA is compared with six representative MOEAs, namely, NSGA-II, MOEA/D-DE, MOEA/D-CMA, IM-MOEA, GDE3, and SPEA2.
The statistical results demonstrate the superiority of GMOEA in solving MOPs with relatively high-dimensional decision variables.

This work demonstrates that the MOEA driven by the GAN is promising in solving MOPs.
Therefore, it deserves further efforts to introduce more efficient generative models.
Besides, the extension of our proposed GMOEA to MOPs with more than three objectives (many-objective optimization problems) is highly desirable.
Moreover, its applications to real-world optimization problems are also meaningful.

\bibliography{Reference}

\begin{thebibliography}{10}
\providecommand{\url}[1]{#1}
\csname url@samestyle\endcsname
\providecommand{\newblock}{\relax}
\providecommand{\bibinfo}[2]{#2}
\providecommand{\BIBentrySTDinterwordspacing}{\spaceskip=0pt\relax}
\providecommand{\BIBentryALTinterwordstretchfactor}{4}
\providecommand{\BIBentryALTinterwordspacing}{\spaceskip=\fontdimen2\font plus
\BIBentryALTinterwordstretchfactor\fontdimen3\font minus
  \fontdimen4\font\relax}
\providecommand{\BIBforeignlanguage}[2]{{%
\expandafter\ifx\csname l@#1\endcsname\relax
\typeout{** WARNING: IEEEtran.bst: No hyphenation pattern has been}%
\typeout{** loaded for the language `#1'. Using the pattern for}%
\typeout{** the default language instead.}%
\else
\language=\csname l@#1\endcsname
\fi
#2}}
\providecommand{\BIBdecl}{\relax}
\BIBdecl

\bibitem{app-network}
N.~Siddique and H.~Adeli, ``Computational intelligence: synergies of fuzzy
  logic,'' in \emph{Neural Networks and Evolutionary Computing}.\hskip 1em plus
  0.5em minus 0.4em\relax John Wiley {\&} Sons, 2013.

\bibitem{liu2018structure}
J.~Liu, M.~Gong, Q.~Miao, X.~Wang, and H.~Li, ``Structure learning for deep
  neural networks based on multiobjective optimization,'' \emph{IEEE
  Transactions on Neural Networks and Learning Systems}, vol.~29, no.~6, pp.
  2450--2463, 2018.

\bibitem{app-build}
W.~Yu, B.~Li, H.~Jia, M.~Zhang, and D.~Wang, ``Application of multi-objective
  genetic algorithm to optimize energy efficiency and thermal comfort in
  building design,'' \emph{Energy and Buildings}, vol.~88, pp. 135--143, 2015.

\bibitem{ferreira2017multi}
P.~V.~R. Ferreira, R.~Paffenroth, A.~M. Wyglinski, T.~M. Hackett, S.~G.
  Bil{\'e}n, R.~C. Reinhart, and D.~J. Mortensen, ``Multi-objective
  reinforcement learning-based deep neural networks for cognitive space
  communications,'' in \emph{Cognitive Communications for Aerospace
  Applications Workshop}.\hskip 1em plus 0.5em minus 0.4em\relax IEEE, 2017,
  pp. 1--8.

\bibitem{deb2014multi}
K.~Deb, ``Multi-objective optimization,'' in \emph{Search Methodologies}.\hskip
  1em plus 0.5em minus 0.4em\relax Springer, 2014, pp. 403--449.

\bibitem{tian2017effectiveness}
Y.~Tian, H.~Wang, X.~Zhang, and Y.~Jin, ``Effectiveness and efficiency of
  non-dominated sorting for evolutionary multi-and many-objective
  optimization,'' \emph{Complex $\&$ Intelligent Systems}, vol.~3, pp.
  247--263, 2017.

\bibitem{PD}
H.~Wang, Y.~Jin, and X.~Yao, ``Diversity assessment in many-objective
  optimization,'' \emph{IEEE Transactions on Cybernetics}, vol.~47, no.~6, pp.
  1510--1522, 2017.

\bibitem{ENS}
X.~Zhang, Y.~Tian, R.~Cheng, and Y.~Jin, ``An efficient approach to
  non-dominated sorting for evolutionary multi-objective optimization,''
  \emph{IEEE Transactions on Evolutionary Computation}, vol.~19, pp. 201--213,
  2015.

\bibitem{RVEA}
R.~Cheng, Y.~Jin, M.~Olhofer, and B.~Sendhoff, ``A reference vector guided
  evolutionary algorithm for many-objective optimization,'' \emph{IEEE
  Transactions on Evolutionary Computation}, vol.~20, pp. 773--791, 2016.

\bibitem{NSGA-II}
K.~Deb, A.~Pratap, S.~Agarwal, and T.~Meyarivan, ``A fast and elitist
  multi-objective genetic algorithm: {NSGA-II},'' \emph{IEEE Transactions on
  Evolutionary Computation}, vol.~6, pp. 182--197, 2002.

\bibitem{SPEA2}
E.~Ziztler, M.~Laumanns, and L.~Thiele, ``{SPEA2}: {I}mproving the strength
  {P}areto evolutionary algorithm for multiobjective optimization,''
  \emph{Evolutionary Methods for Design, Optimization, and Control}, pp.
  95--100, 2002.

\bibitem{MOEAD}
Q.~Zhang and H.~Li, ``{MOEA/D}: {A} multiobjective evolutionary algorithm based
  on decomposition,'' \emph{IEEE Transactions on Evolutionary Computation},
  vol.~11, pp. 712--731, 2007.

\bibitem{MOEADDE}
H.~Li and Q.~Zhang, ``Multiobjective optimization problems with complicated
  {P}areto sets, {MOEA/D} and {NSGA-II},'' \emph{IEEE Transactions on
  Evolutionary Computation}, vol.~13, pp. 284--302, 2009.

\bibitem{SMSEMOA}
N.~Beume, B.~Naujoks, and M.~Emmerich, ``{SMS-EMOA}: Multiobjective selection
  based on dominated hypervolume,'' \emph{European Journal of Operational
  Research}, vol. 181, no.~3, pp. 1653--1669, 2007.

\bibitem{IBEA}
E.~Zitzler and S.~K\"{u}nzli, ``Indicator-based selection in multiobjective
  search,'' in \emph{International Conference on Parallel Problem Solving from
  Nature}.\hskip 1em plus 0.5em minus 0.4em\relax Springer, 2004, pp. 832--842.

\bibitem{GDE3}
L.~M. Antonio and C.~A.~C. Coello, ``Use of cooperative coevolution for solving
  large scale multiobjective optimization problems,'' in \emph{IEEE Congress on
  Evolutionary Computation}.\hskip 1em plus 0.5em minus 0.4em\relax IEEE, 2013,
  pp. 2758--2765.

\bibitem{knowles2000m}
J.~D. Knowles and D.~W. Corne, ``{M-PAES}: A memetic algorithm for
  multiobjective optimization,'' in \emph{IEEE Congress on Evolutionary
  Computation}.\hskip 1em plus 0.5em minus 0.4em\relax IEEE, 2000, pp.
  325--332.

\bibitem{praditwong2006new}
K.~Praditwong and X.~Yao, ``A new multi-objective evolutionary optimisation
  algorithm: The two-archive algorithm,'' in \emph{International Conference on
  Computational Intelligence and Security}.\hskip 1em plus 0.5em minus
  0.4em\relax IEEE, 2006, pp. 286--291.

\bibitem{eiben2015evolutionary}
A.~E. Eiben and J.~Smith, ``From evolutionary computation to the evolution of
  things,'' \emph{Nature}, vol. 521, no. 7553, p. 476, 2015.

\bibitem{PM}
K.~Deb and M.~Goyal, ``A combined genetic adaptive search (geneas) for
  engineering design,'' \emph{Computer Science and Informatics}, vol.~26, pp.
  30--45, 1996.

\bibitem{IM-MOEA}
R.~Cheng, Y.~Jin, K.~Narukawa, and B.~Sendhoff, ``A multiobjective evolutionary
  algorithm using {G}aussian process-based inverse modeling,'' \emph{IEEE
  Transactions on Evolutionary Computation}, vol.~19, pp. 838--856, 2015.

\bibitem{MBEA}
R.~Cheng, C.~He, Y.~Jin, and X.~Yao, ``Model-based evolutionary algorithms: a
  short survey,'' \emph{Complex \& Intelligent Systems}, pp. 1--10, 2018.

\bibitem{zhang2011evolutionary}
J.~Zhang, Z.-h. Zhan, Y.~Lin, N.~Chen, Y.-j. Gong, J.-h. Zhong, H.~S. Chung,
  Y.~Li, and Y.-h. Shi, ``Evolutionary computation meets machine learning: A
  survey,'' \emph{IEEE Computational Intelligence Magazine}, vol.~6, no.~4, pp.
  68--75, 2011.

\bibitem{Jin2000On}
Y.~Jin, M.~Olhofer, and B.~Sendhoff, ``On evolutionary optimization with
  approximate fitness functions,'' in \emph{Genetic and Evolutionary
  Computation Conference}, 2000, pp. 786--793.

\bibitem{jin2009systems}
Y.~Jin and B.~Sendhoff, ``A systems approach to evolutionary multiobjective
  structural optimization and beyond,'' \emph{IEEE Computational Intelligence
  Magazine}, vol.~4, no.~3, pp. 62--76, 2009.

\bibitem{jin2011review}
Y.~Jin, ``Surrogate-assisted evolutionary computation: Recent advances and
  future challenges,'' \emph{Swarm and Evolutionary Computation}, vol.~1,
  no.~2, pp. 61--70, 2011.

\bibitem{SA2017}
R.~Allmendinger, M.~T.~M.~Emmerich, J.~Hakanen, Y.~J. Jin, and E.~Rigoni,
  ``Surrogate-assisted multicriteria optimization: Complexities, prospective
  solutions, and business case,'' \emph{Journal of Multi-Criteria Decision
  Analysis}, vol.~24, no. 1-2, pp. 5--24, 2017.

\bibitem{SMS-EGO}
W.~Ponweiser, T.~Wagner, D.~Biermann, and M.~Vincze, ``Multiobjective
  optimization on a limited budget of evaluations using model-assisted
  {$\mathcal{S}$}-{M}etric selection,'' in \emph{International Conference on
  Parallel Problem Solving from Nature}.\hskip 1em plus 0.5em minus 0.4em\relax
  Springer, 2008, pp. 784--794.

\bibitem{seah2012pareto}
C.-W. Seah, Y.-S. Ong, I.~W. Tsang, and S.~Jiang, ``Pareto rank learning in
  multi-objective evolutionary algorithms,'' in \emph{IEEE Congress on
  Evolutionary Computation}.\hskip 1em plus 0.5em minus 0.4em\relax IEEE, 2012.

\bibitem{GP}
J.~R. Koza, ``Genetic programming as a means for programming computers by
  natural selection,'' \emph{Statistics and Computing}, vol.~4, no.~2, pp.
  87--112, 1994.

\bibitem{MOEADEGO}
Q.~Zhang, W.~Liu, E.~Tsang, and B.~Virginas, ``Expensive multiobjective
  optimization by {MOEA/D} with {G}aussian process model,'' \emph{IEEE
  Transactions on Evolutionary Computation}, vol.~14, pp. 456--474, 2010.

\bibitem{ParetoSVM}
I.~Loshchilov, M.~Schoenauer, and M.~Sebag, ``A mono surrogate for
  multiobjective optimization,'' in \emph{Annual Conference on Genetic and
  Evolutionary Computation}.\hskip 1em plus 0.5em minus 0.4em\relax ACM, 2010,
  pp. 471--478.

\bibitem{lu2012classification}
X.~Lu and K.~Tang, ``Classification-and regression-assisted differential
  evolution for computationally expensive problems,'' \emph{Journal of Computer
  Science and Technology}, vol.~27, no.~5, pp. 1024--1034, 2012.

\bibitem{bhatt2015novel}
K.~S. Bhattacharjee and T.~Ray, ``A novel constraint handling strategy for
  expensive optimization problems,'' in \emph{World Congress on Structural and
  Multidisciplinary Optimization}, 2015.

\bibitem{CPSMOEA}
J.~Zhang, A.~Zhou, and G.~Zhang, ``A classification and {P}areto domination
  based multiobjective evolutionary algorithm,'' in \emph{IEEE Congress on
  Evolutionary Computation}.\hskip 1em plus 0.5em minus 0.4em\relax IEEE, 2015,
  pp. 2883--2890.

\bibitem{KNN}
A.~K. Jain and R.~C. Dubes, ``Algorithms for clustering data,''
  \emph{Technometrics}, vol.~32, no.~2, pp. 227--229, 1988.

\bibitem{zhang2018preselection}
J.~Zhang, A.~Zhou, K.~Tang, and G.~Zhang, ``Preselection via classification: A
  case study on evolutionary multiobjective optimization,'' \emph{Information
  Sciences}, vol. 465, pp. 388--403, 2018.

\bibitem{svozil1997introduction}
D.~Svozil, V.~Kvasnicka, and J.~Pospichal, ``Introduction to multi-layer
  feed-forward neural networks,'' \emph{Chemometrics and Intelligent Laboratory
  Systems}, vol.~39, no.~1, pp. 43--62, 1997.

\bibitem{CSEA}
L.~Pan, C.~He, Y.~Tian, H.~Wang, X.~Zhang, and Y.~Jin, ``A classification based
  surrogate{-}assisted evolutionary algorithm for expensive many{-}objective
  optimization,'' \emph{IEEE Transactions on Evolutionary Computation}, 2018.

\bibitem{EDA}
P.~Larra{\~n}aga and J.~A. Lozano, \emph{Estimation of distribution algorithms:
  A new tool for evolutionary computation}.\hskip 1em plus 0.5em minus
  0.4em\relax Springer Science \& Business Media, 2001, vol.~2.

\bibitem{giagkiozis2014pareto}
I.~Giagkiozis and P.~J. Fleming, ``{Pareto} front estimation for decision
  making,'' \emph{Evolutionary Computation}, vol.~22, no.~4, pp. 651--678,
  2014.

\bibitem{karshenas2013multiobjective}
H.~Karshenas, R.~Santana, C.~Bielza, and P.~Larranaga, ``Multiobjective
  estimation of distribution algorithm based on joint modeling of objectives
  and variables,'' \emph{IEEE Transactions on Evolutionary Computation},
  vol.~18, no.~4, pp. 519--542, 2013.

\bibitem{eda-initial}
P.~Larranaga and J.~A. Lozano, \emph{Estimation of Distribution Algorithms: A
  New Tool for Evolutionary Computation}.\hskip 1em plus 0.5em minus
  0.4em\relax Springer Science \& Business Media, 2002.

\bibitem{sun2018improved}
Y.~Sun, G.~G. Yen, and Z.~Yi, ``Improved regularity model-based {EDA} for
  many-objective optimization,'' \emph{IEEE Transactions on Evolutionary
  Computation}, vol.~22, no.~5, pp. 662--678, 2018.

\bibitem{BMOA}
M.~Laumanns and J.~Ocenasek, ``Bayesian optimization algorithms for
  multi-objective optimization,'' in \emph{International Conference on Parallel
  Problem Solving from Nature}.\hskip 1em plus 0.5em minus 0.4em\relax
  Springer, 2002, pp. 298--307.

\bibitem{MIDEA}
P.~A. Bosman and D.~Thierens, ``Multi-objective optimization with the naive
  $\mathbb{M}${ID}$\mathbb{E}${A},'' in \emph{Towards a New Evolutionary
  Computation}.\hskip 1em plus 0.5em minus 0.4em\relax Springer, 2006, pp.
  123--157.

\bibitem{mBOA}
J.~Ocenasek, S.~Kern, N.~Hansen, and P.~Koumoutsakos, ``A mixed {B}ayesian
  optimization algorithm with variance adaptation,'' in \emph{International
  Conference on Parallel Problem Solving from Nature}.\hskip 1em plus 0.5em
  minus 0.4em\relax Springer, 2004, pp. 352--361.

\bibitem{RM-MEDA}
Q.~Zhang, A.~Zhou, and Y.~Jin, ``{RM-MEDA}: A regularity model-based
  multiobjective estimation of distribution algorithm,'' \emph{IEEE
  Transactions on Evolutionary Computation}, vol.~12, pp. 41--63, 2008.

\bibitem{MOEADCMA}
H.~Li, Q.~Zhang, and J.~Deng, ``Biased multiobjective optimization and
  decomposition algorithm,'' \emph{IEEE Transactions on Cybernetics}, vol.~47,
  no.~1, pp. 52--66, 2017.

\bibitem{loshchilov2013cma}
I.~Loshchilov, ``{CMA-ES} with restarts for solving {CEC} 2013 benchmark
  problems,'' in \emph{IEEE Congress on Evolutionary Computation}.\hskip 1em
  plus 0.5em minus 0.4em\relax Ieee, 2013, pp. 369--376.

\bibitem{hernandez2007pareto}
A.~G. Hern{\'a}ndez-D{\'\i}az, L.~V. Santana-Quintero, C.~A. Coello~Coello, and
  J.~Molina, ``Pareto-adaptive $\varepsilon$-dominance,'' \emph{Evolutionary
  Computation}, vol.~15, no.~4, pp. 493--517, 2007.

\bibitem{martinez2014using}
S.~Z. Mart{\'\i}nez, V.~A.~S. Hern{\'a}ndez, H.~Aguirre, K.~Tanaka, and
  C.~A.~C. Coello, ``Using a family of curves to approximate the {P}areto front
  of a multi-objective optimization problem,'' in \emph{International
  Conference on Parallel Problem Solving from Nature}.\hskip 1em plus 0.5em
  minus 0.4em\relax Springer, 2014, pp. 682--691.

\bibitem{cd}
L.~Parsons, E.~Haque, and H.~Liu, ``Subspace clustering for high dimensional
  data: a review,'' \emph{ACM SIGKDD Explorations Newsletter}, vol.~6, no.~1,
  pp. 90--105, 2004.

\bibitem{wang2015memetic}
H.~Wang, L.~Jiao, R.~Shang, S.~He, and F.~Liu, ``A memetic optimization
  strategy based on dimension reduction in decision space,'' \emph{Evolutionary
  Computation}, vol.~23, no.~1, pp. 69--100, 2015.

\bibitem{GAN}
I.~Goodfellow, J.~Pouget-Abadie, M.~Mirza, B.~Xu, D.~Warde-Farley, S.~Ozair,
  A.~Courville, and Y.~Bengio, ``Generative adversarial nets,'' in
  \emph{Advances in Neural Information Processing Systems}, 2014, pp.
  2672--2680.

\bibitem{radford2015unsupervised}
A.~Radford, L.~Metz, and S.~Chintala, ``Unsupervised representation learning
  with deep convolutional generative adversarial networks,'' \emph{arXiv
  preprint arXiv:1511.06434}, 2015.

\bibitem{ledig2017photo}
C.~Ledig, L.~Theis, F.~Husz{\'a}r \emph{et~al.}, ``Photo-realistic single image
  super-resolution using a generative adversarial network,'' in \emph{CVPR},
  vol.~2, no.~3, 2017, p.~4.

\bibitem{kingma2014adam}
D.~P. Kingma and J.~Ba, ``Adam: A method for stochastic optimization,''
  \emph{arXiv preprint arXiv:1412.6980}, 2014.

\bibitem{SPEA}
E.~Zitzler and L.~Thiele, ``Multiobjective evolutionary algorithms: a
  comparative case study and the strength {P}areto approach,'' \emph{IEEE
  Transactions on Evolutionary Computation}, vol.~3, pp. 257--271, 1999.

\bibitem{DL}
I.~Goodfellow, Y.~Bengio, A.~Courville, and Y.~Bengio, \emph{Deep
  Learning}.\hskip 1em plus 0.5em minus 0.4em\relax MIT press Cambridge, 2016,
  vol.~1.

\bibitem{ANN}
M.~Leshno, V.~Y. Lin, A.~Pinkus, and S.~Schocken, ``Multilayer feedforward
  networks with a nonpolynomial activation function can approximate any
  function,'' \emph{Neural Networks}, vol.~6, no.~6, pp. 861--867, 1993.

\bibitem{zhang2018computational}
X.~Zhang, B.~Ding, R.~Cheng, S.~C. Dixon, and Y.~Lu, ``Computational
  intelligence-assisted understanding of nature-inspired superhydrophobic
  behavior,'' \emph{Advanced Science}, vol.~5, no.~1, p. 1700520, 2018.

\bibitem{balakrishnan2014continuous}
N.~Balakrishnan, ``Continuous multivariate distributions,'' \emph{Wiley
  StatsRef: Statistics Reference Online}, 2014.

\bibitem{VAE}
D.~P. Kingma and M.~Welling, ``Auto-encoding variational bayes,'' \emph{arXiv
  preprint arXiv:1312.6114}, 2013.

\bibitem{arjovsky2017wasserstein}
M.~Arjovsky, S.~Chintala, and L.~Bottou, ``Wasserstein {GAN},'' \emph{arXiv
  preprint arXiv:1701.07875}, 2017.

\bibitem{haynes2013wilcoxon}
W.~Haynes, ``Wilcoxon rank sum test,'' in \emph{Encyclopedia of Systems
  Biology}.\hskip 1em plus 0.5em minus 0.4em\relax Springer, 2013, pp.
  2354--2355.

\bibitem{PlatEMO}
Y.~Tian, R.~Cheng, X.~Zhang, and Y.~Jin, ``{PlatEMO}: A {MATLAB} platform for
  evolutionary multi-objective optimization,'' \emph{IEEE Computational
  Intelligence Magazine}, vol.~12, pp. 73--87, 2017.

\bibitem{review-book}
K.~Deb, \emph{Multi-objective optimization using evolutionary
  algorithms}.\hskip 1em plus 0.5em minus 0.4em\relax John Wiley \& Sons, 2001,
  vol.~16.

\bibitem{DE}
R.~Storn and K.~Price, ``Differential evolution--a simple and efficient
  heuristic for global optimization over continuous spaces,'' \emph{Journal of
  Global Optimization}, vol.~11, no.~4, pp. 341--359, 1997.

\bibitem{Adam}
D.~P. Kingma and J.~Ba, ``{Adam}: A method for stochastic optimization,''
  \emph{arXiv preprint arXiv:1412.6980}, 2014.

\bibitem{IGD}
A.~Zhou, Y.~Jin, Q.~Zhang, B.~Sendhoff, and E.~Tsang, ``Combining model-based
  and genetics-based offspring generation for multi-objective optimization
  using a convergence criterion,'' in \emph{IEEE Congress on Evolutionary
  Computation}, 2006, pp. 892--899.

\bibitem{RPgeneration}
C.~He, L.~Pan, H.~Xu, Y.~Tian, and X.~Zhang, ``An improved reference point
  sampling method on {P}areto optimal front,'' in \emph{IEEE Congress on
  Evolutionary Computation}.\hskip 1em plus 0.5em minus 0.4em\relax IEEE, 2016,
  pp. 5230--5237.

\bibitem{HV}
L.~While, P.~Hingston, L.~Barone, and S.~Huband, ``A faster algorithm for
  calculating hypervolume,'' \emph{IEEE Transactions on Evolutionary
  Computation}, vol.~10, pp. 29--38, 2006.

\bibitem{pytorch}
N.~Ketkar, \emph{Introduction to pytorch}.\hskip 1em plus 0.5em minus
  0.4em\relax Springer, 2017.

\end{thebibliography}
\bibliographystyle{IEEEtran}

\begin{IEEEbiography}[{\includegraphics[width=1in,height=1.2in,clip,keepaspectratio]{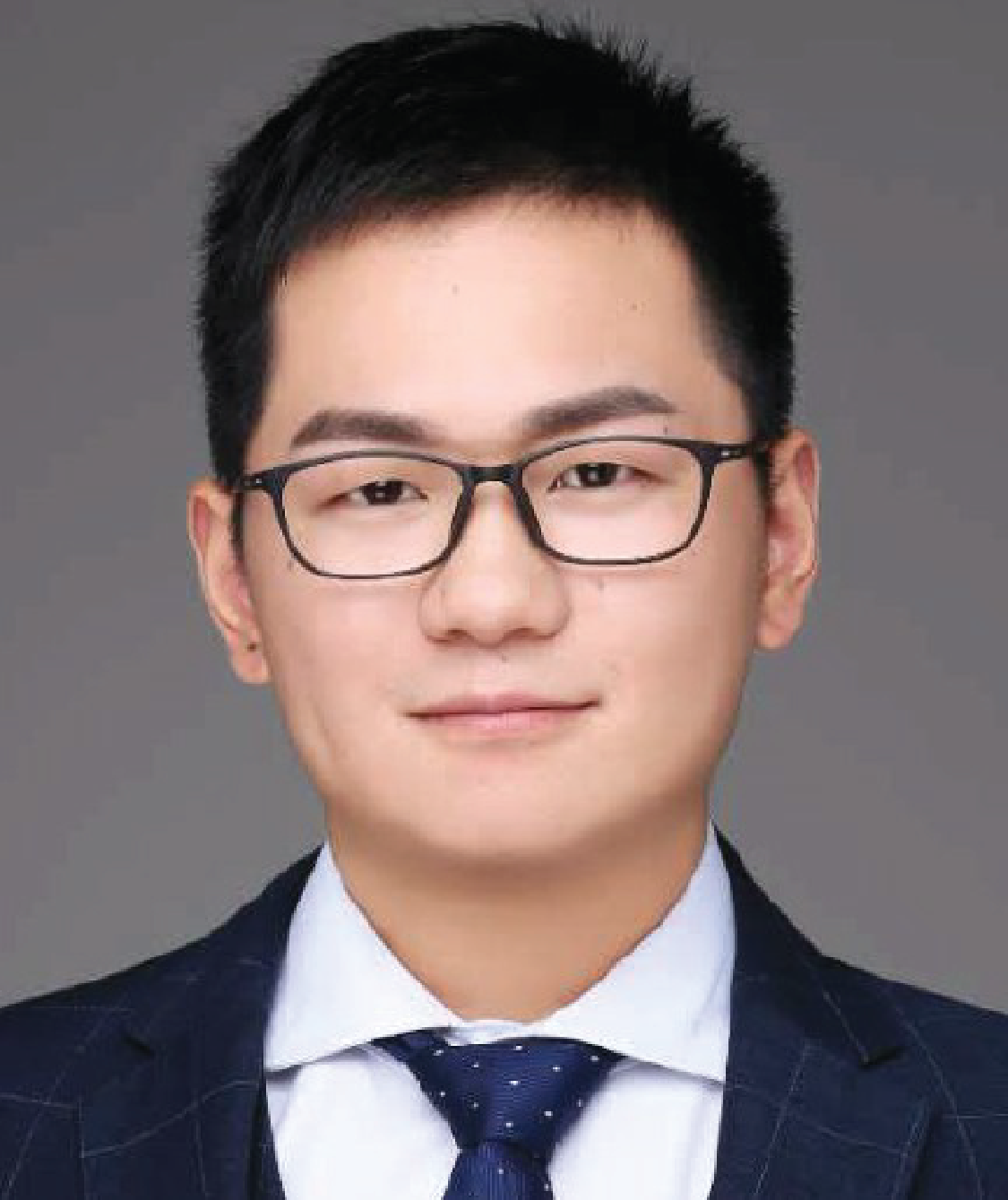}}]{Cheng He} (M'2019)
received the B.Eng. degree from the Wuhan University of Science and Technology, Wuhan, China, in 2012, and the Ph.D. degree from the Huazhong University of Science and Technology, Wuhan, China, in 2018.

He is currently a Postdoctoral Research Fellow with the Department of Computer Science and Engineering, Southern University of Science and Technology, Shenzhen, China. His current research interests include model-based evolutionary algorithms, multiobjective optimization, large-scale optimization, deep learning, and their applications. He is a recipient of the SUSTech Presidential Outstanding Postdoctoral Award from Southern University of Science and Technology, and a member of IEEE Task Force on Data-Driven Evolutionary Optimization of Expensive Problems.
\end{IEEEbiography}

\begin{IEEEbiography}[{\includegraphics[width=1in,height=1.2in,clip,keepaspectratio]{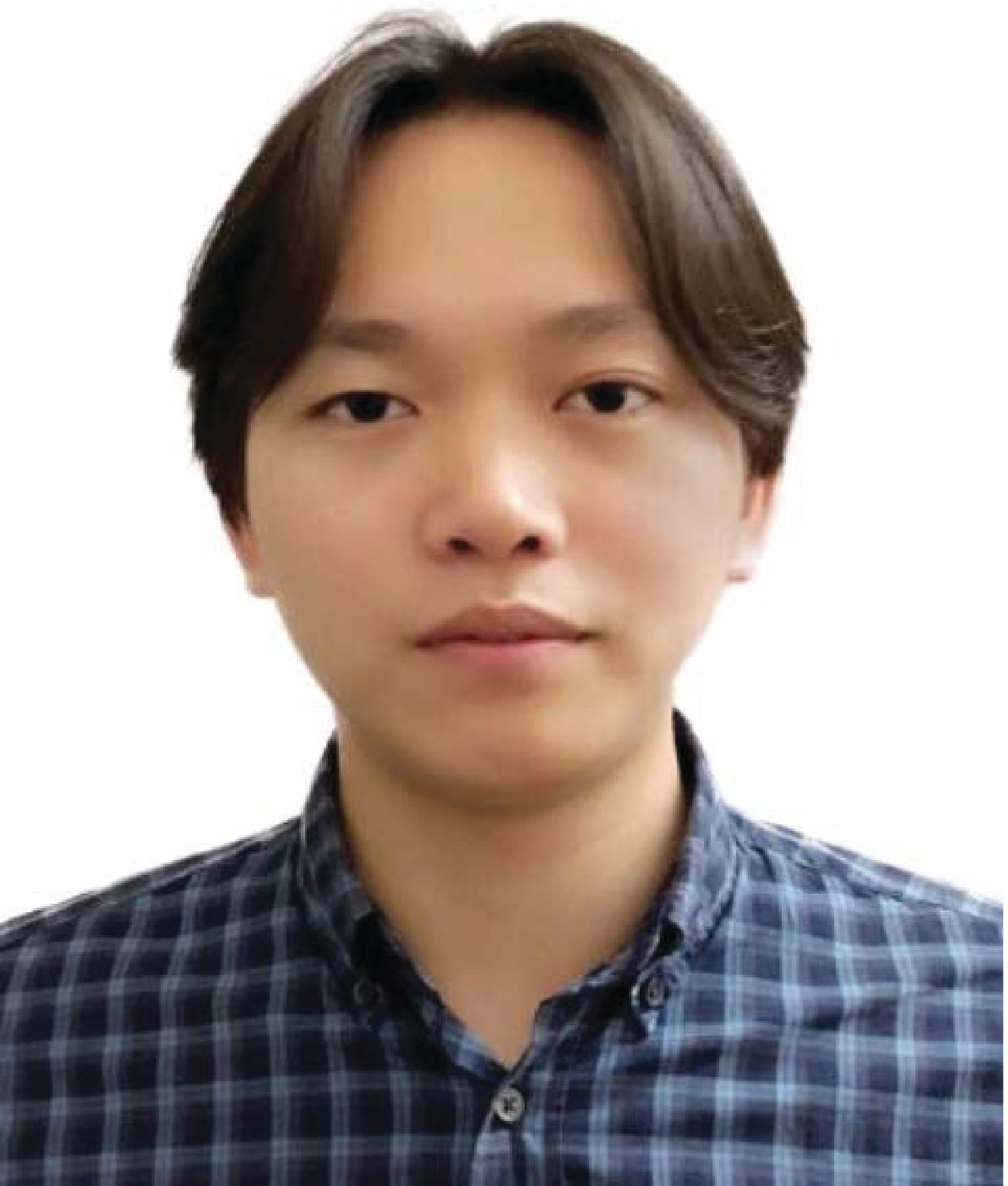}}]{Shihua Huang}
received the B.Eng. degree from the Northeastern University, Shenyang, China, in 2018.

Currently, he is a research assistant with the Department of Computer Science and Engineering, Southern University of Science and Technology, Shenzhen, China.
His current research interests include deep learning, multiobjective optimization, and their application.
\end{IEEEbiography}

\begin{IEEEbiography}[{\includegraphics[width=1in,height=1.2in,clip,keepaspectratio]{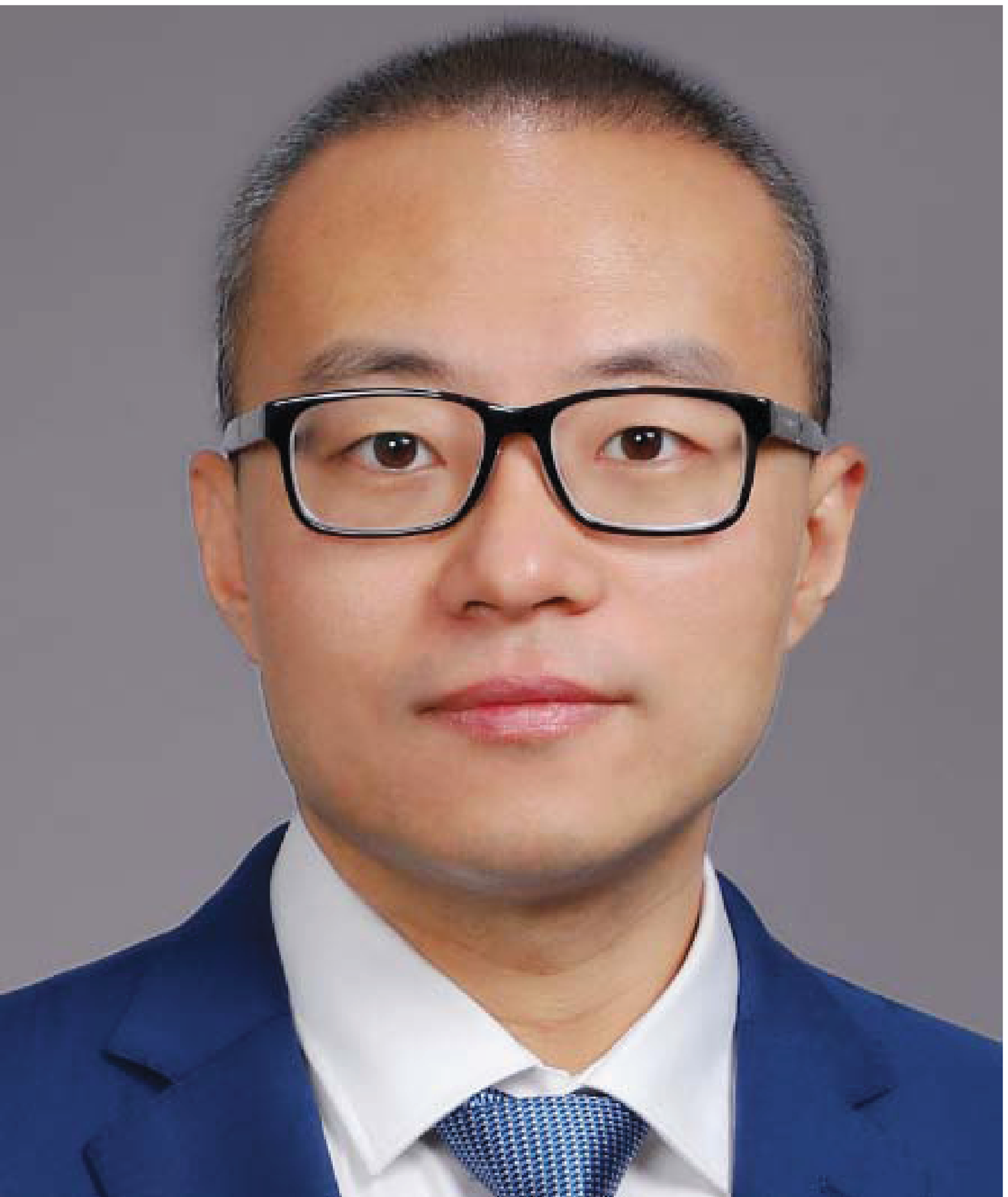}}]{Ran Cheng} (M'2016) received the B.Sc. degree from the Northeastern University, Shenyang, China, in 2010, and the Ph.D. degree from the University of Surrey, Guildford, U.K., in 2016.

He is currently an Assistant Professor with the Department of Computer Science and Engineering, Southern University of Science and Technology, Shenzhen, China. His current research interests include evolutionary multiobjective optimization, model-based evolutionary algorithms, large-scale optimization, swarm intelligence, and deep learning.
He is the founding chair of IEEE Symposium on Model Based Evolutionary Algorithms (IEEE MBEA).
 He is currently an Associate Editor of the IEEE Transactions on Artificial Intelligence and the IEEE Access.
 He is the recipient of the 2018 IEEE Transactions on Evolutionary Computation Outstanding Paper Award, the 2019 IEEE Computational Intelligence Society (CIS) Outstanding Ph.D. Dissertation Award, and the 2020 IEEE Computational Intelligence Magazine Outstanding Paper Award.
\end{IEEEbiography}

\begin{IEEEbiography}[{\includegraphics[width=1in,height=1.2in,clip,keepaspectratio]{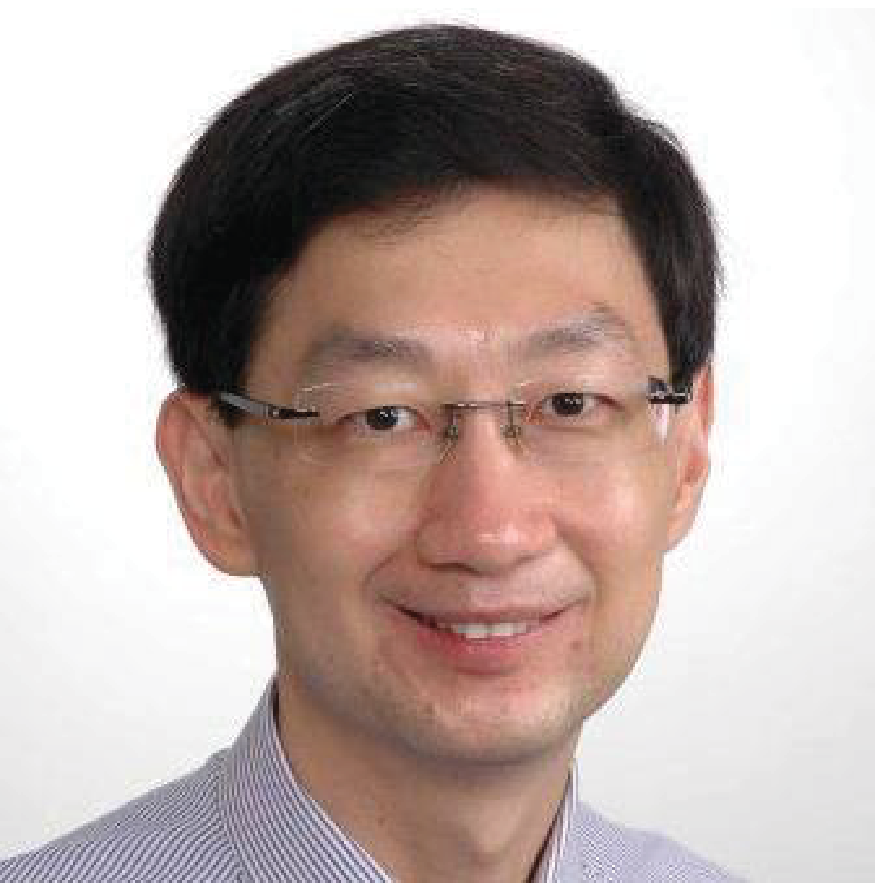}}]{Kay Chen Tan} (SM'08-F'14) received the B.Eng. degree (First Class Hons.) and the Ph.D. degree from the University of Glasgow, U.K., in 1994 and 1997, respectively.
He is a currently full Professor with the Department of Computer Science, City University of Hong Kong.
He is the Editor-in-Chief of IEEE Transactions on Evolutionary Computation, was the EiC of IEEE Computational Intelligence Magazine (2010-2013), and currently serves on the Editorial Board member of 20+ journals.
He is a Fellow of IEEE, an elected AdCom member of IEEE Computational Intelligence Society (2017-2019).
He has published 200+ refereed articles and 6 books.
\end{IEEEbiography}

\begin{IEEEbiography}[{\includegraphics[width=1in,height=1.2in,clip,keepaspectratio]{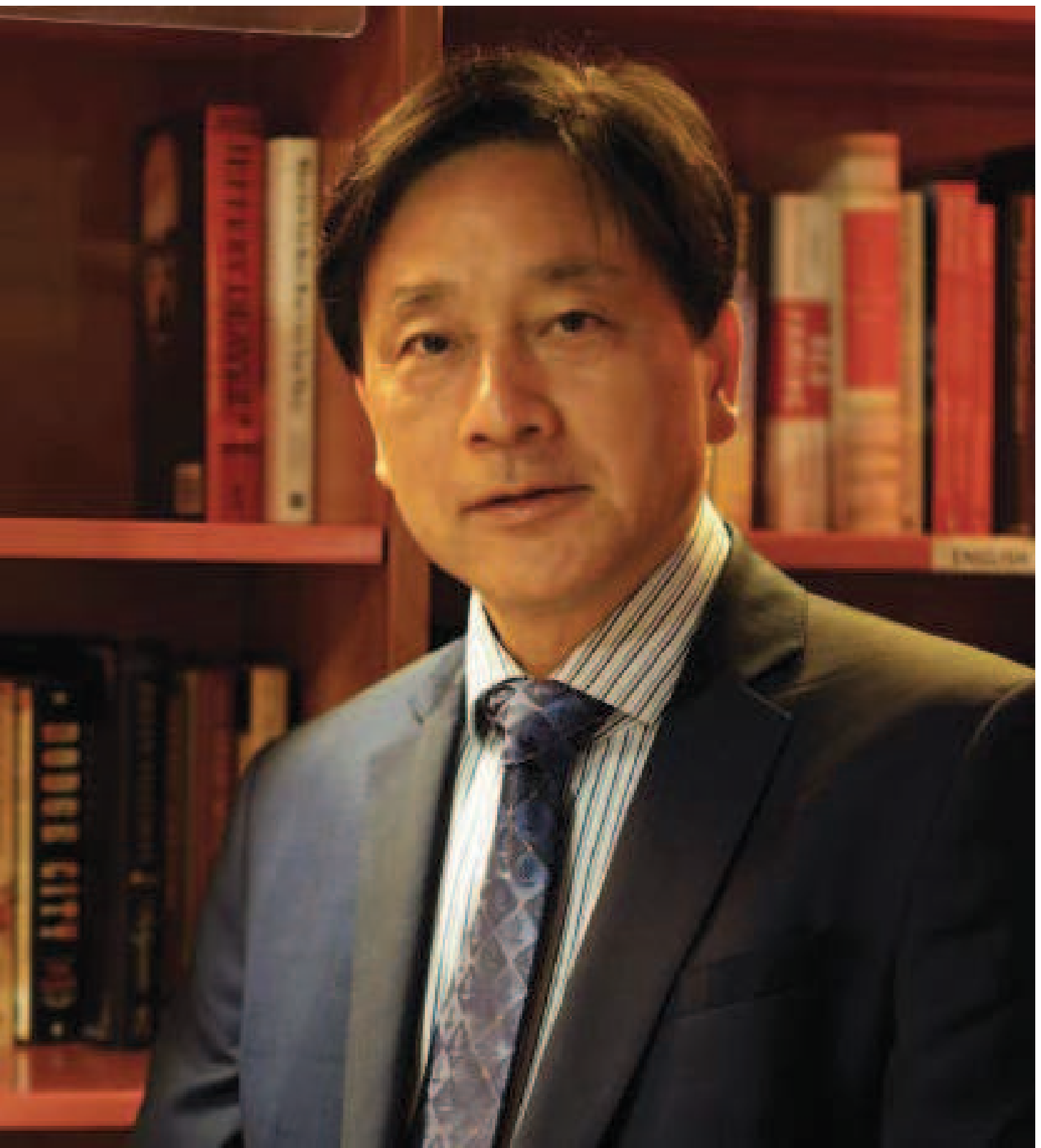}}]{Yaochu Jin} (M'98-SM'02-F'16) received the B.Sc., M.Sc., and Ph.D. degrees from Zhejiang University, Hangzhou, China, in 1988, 1991, and 1996, respectively, and the Dr.-Ing. degree from Ruhr University Bochum, Germany, in 2001.

He is currently a Professor in Computational Intelligence, Department of Computer Science, University of Surrey, Guildford, U.K., where he heads the Nature Inspired Computing and Engineering Group. He is also a Finland Distinguished Professor funded by the Finnish Funding Agency for Innovation (Tekes), Finland and a Changjiang Distinguished Visiting Professor appointed by the Ministry of Education, China. His research interests lie primarily in the cross-disciplinary areas of computational intelligence, computational neuroscience, and computational systems biology. He is also particularly interested in the application of nature-inspired algorithms to solving real-world optimization, learning and self-organization problems. He has (co)authored over 300 peer-reviewed journal and conference papers and been granted eight patents on evolutionary optimization.

Dr Jin is the Editor-in-Chief of the IEEE TRANSACTIONS ON COGNITIVE AND DEVELOPMENTAL SYSTEMS and Complex \& Intelligent Systems. He is an IEEE Distinguished Lecturer (2017-2019) and was the Vice President for Technical Activities of the IEEE Computational Intelligence Society (2014-2015). He is a recipient of the 2015 and 2017 IEEE Computational Intelligence Magazine Outstanding Paper Award, and the 2018 IEEE Transactions on Evolutionary Computation Outstanding Paper Award. He is a Fellow of IEEE.
\end{IEEEbiography}

\end{document}